\documentclass[aip,cha,amsmath,amssymb,reprint,floatfix,numbers]{revtex4-1}

\usepackage[T1]{fontenc}
\usepackage[latin9]{inputenc}
\usepackage{color}
\usepackage{amstext}
\usepackage{graphicx}
\usepackage{hyperref}

\makeatletter
\usepackage{dynlearn}
\usepackage{multirow}

\newcommand{\Eps}{\epsilon}
\newcommand{\altx}{x^\prime}

\newcommand{\keM}{kernel \eM}
\newcommand{\KeM}{Kernel \EM}
\newcommand{\keMs}{kernel \eMs}
\newcommand{\KeMs}{Kernel \EMs}

\makeatother

\usepackage{babel}

\begin{document}

\title{Discovering Causal Structure with\\ Reproducing-Kernel Hilbert Space \EMs}

\author{Nicolas Brodu}
\email{nicolas.brodu@inria.fr}
\affiliation{Geostat Team - Geometry and Statistics in Acquisition Data, INRIA
Bordeaux Sud Ouest~~\\
 200 rue de la Vieille Tour, 33405 Talence Cedex, France}

\author{James P. Crutchfield}
\email{chaos@ucdavis.edu}
\affiliation{Complexity Sciences Center and Department of Physics and
Astronomy, University of California at Davis, One Shields Avenue, Davis, CA
95616}

\date{\today}

\begin{abstract}
We merge computational mechanics' definition of causal states
(predictively-equivalent histories) with reproducing-kernel Hilbert space
(RKHS) representation inference. The result is a widely-applicable method that
infers causal structure directly from observations of a system's behaviors
whether they are over discrete or continuous events or time. A structural
representation---a finite- or infinite-state \keM---is extracted by a
reduced-dimension transform that gives an efficient representation of causal
states and their topology. In this way, the system dynamics are represented by
a stochastic (ordinary or partial) differential equation that acts on causal
states. We introduce an algorithm to estimate the associated evolution
operator. Paralleling the Fokker-Plank equation, it efficiently evolves
causal-state distributions and makes predictions in the original data space via
an RKHS functional mapping. We demonstrate these techniques, together with
their predictive abilities, on discrete-time, discrete-value infinite
Markov-order processes generated by finite-state hidden Markov models with (i)
finite or (ii) uncountably-infinite causal states and (iii) continuous-time,
continuous-value processes generated by thermally-driven chaotic flows. The
method robustly estimates causal structure in the presence of varying external
and measurement noise levels and for very high dimensional data.
\end{abstract}

\preprint{\arxiv{2020.XXXXX}}

\maketitle
\setstretch{1.1}


\begin{quotation}
Computational mechanics is a mathematical framework for pattern discovery that
describes how information is stored, structured, and transformed in a physical
process. Its constructive application to observed data has been demonstrated
for some time. Until now, though, success was limited by the need to strongly
discretize observations or discover state-space generating partitions for
correct symbolic dynamics. Exploiting modern machine-learning foundations in
functional analysis, we broadly extend computational mechanics to inferring
models of many distinct classes of structured process, going beyond
fully-discrete data to processes with continuous data and those measured by
heterogeneous instruments. Equations of motion for the evolution of process
states can also be reconstructed from data. The method successfully recovers a
process's causal states and its dynamics in both discrete and continuous cases,
including the recovery of noisy, high-dimensional chaotic attractors.
\end{quotation}

\section{Introduction}
\label{sec:Intro}

At root, the physical sciences and engineering turn on successfully modeling
the behavior of a physical system from observations. Noting that they have been
successful in this is an understatement, at best. That success begs a deep
question, though---one requiring careful reflection. How, starting only from
measurements, does one construct a model for behavioral evolution consistent
with the given data?

Not surprisingly, dynamical-model inference has been tackled via a variety of
theoretical frameworks. However, most approaches make strong assumptions about
the internal organization of the data-generating process: Fourier analysis
assumes a collection of exactly-periodic oscillators; Laplace transforms, a
collection of exponential relaxation processes; feed-forward neural networks,
linearly-coupled threshold units. We are all familiar with the results: A
method is very successful when the data generator is in the assumed model
class. The question begged in this is that one must know something---for some
process classes, quite a lot--- about the data before starting a successful
analysis. And, if the wrong model class is assumed, this strategy tells one
vanishingly little or nothing about how to find the correct one. This form of
model inference is what we call \emph{pattern recognition}: Does the assumed
model class fairly represent the data and, in particular, the generator's
organization?

At some level of abstraction, one cannot escape these issues. Yet, the
assumptions required by our framework are sufficiently permissive that we can
attempt \emph{pattern discovery}, in contrast with pattern recognition. Can we
learn new structures, not assumed? Can we extract efficient representations of
the data and their evolution? Do these yield optimal predictions?

Framed this way, it is not surprising that contemporary model inference is an
active research area---one in which success is handsomely rewarded. Indeed, it
is becoming increasingly important in our current era of massive data sets and
cloud computing.

The following starts from and then extends \emph{computational mechanics}
\citep{Crut88a,Crut12a}---a mathematical framework that lays out the foundations
for pattern discovery. The core idea is to statistically describe a system's
evolution in terms of causally-equivalent states---effective states, built only
from given data, that lead to the same consequences in a system's future. Its
theorems state that causal states and their transition dynamic can be
reconstructed from data, giving the minimal, optimally-predictive
model.\citep{Shal98a} Constructively, for discrete-event, discrete-time or
continuous-time processes computational mechanics delineates the minimal causal
representations---their \emph{\eMs}.\cite{Marz20a,Marz17b} This leaves open,
for example, \eMs for the rather large and important class of continuous-value,
continuous-time processes and also related spatiotemporal processes. The
practical parallel to these remaining challenges is that currently there is no
single algorithm that reconstructs \eMs from data in a reasonable time and
without substantial discrete approximations.\citep{Shal02a,Goerg13,rupe2019disco}

Thus, one goal of the following is to extend computational mechanics beyond its
current domains to continuous time and arbitrary data. This results in a new
class of inference algorithm for \emph{\keMs} that, notably, can be less
resource-demanding than their predecessors. At the very least, the algorithms
work with a set of weaker assumptions about the given data, extending
computational mechanics' structural representations to previously inaccessible
applications with continuous data.

One of the primary motivations for using \eM representations in the first place
is not modeling. Rather, once in hand, their mathematical properties allow
direct and efficient estimation of a range of statistical,
information-theoretic, and thermodynamic properties of a process. This latter
benefit, however, is not the focus of the following; rather those goals are the
target of sequels.

In the new perspective, \keMs are analogous to models associated with Langevin
dynamics, that act on a set of state variables representing system
configurations. The new continuous \keMs can also be written in the form of a
stochastic differential equation (SDE) that acts instead on the
predictively-equivalent causal states. Similarly, a Fokker-Planck equation
describing the evolution of a distribution over causal states can be defined.
It is then used to infer the evolution from an initial probability distribution
over the model's internal states (be it a real distribution induced by
uncertainty or a delta function), which is then used to make predictions in the
original data space. 

The next Section recalls the minimal set of computational mechanics required
for the full development. Section \ref{subsec:states_measures_rkhs} establishes
that the space of causal states has a natural metric and a well-defined measure
using \emph{reproducing kernel Hilbert spaces}. This broadly extends
computational mechanics to many kinds of process, from discrete-value and -time
to continuous-value and -time and everywhere in between, including
spatiotemporal and network dynamical systems. Section
\ref{sec:Dynamics-of-causal-states} then presents the evolution equations and
discusses how to infer from empirical data the evolution operator for
system-state distributions. Section \ref{sec:Validation-Examples} demonstrates
how to apply the algorithm to discover optimal models from realizations of
discrete-time, discrete-value infinite Markov-order process generated by
finite-state hidden Markov models of varying complexity and continuous-time,
continuous-value processes generated by thermally-driven deterministic chaotic
flows.

\section{Computational Mechanics: A synopsis}

We first describe the main objects of study in computational
mechanics---stochastic processes. Then we define the effective or causal states
and their transition dynamics directly in terms of predicting a process'
realizations.

\newcommand{\Borel}{\mathcal{B}}
\newcommand{\Xvals}{\mathcal{X}}
\newcommand{\Yvals}{\mathcal{Y}}
\newcommand{\MSym}{\MeasSymbol}
\newcommand{\msym}{\meassymbol}
\newcommand{\MSs}{\MeasSymbols}
\newcommand{\filtSet}{\mathcal{T}}

\subsection{Processes}
\label{subsec:Processes}

Though targeted to discover a stochastic process' intrinsic structure,
computational mechanics takes the task of predicting a process as its starting
point. To describe how a process' structure is discovered within a purely
predictive framework, we introduce the following setup.

Consider a \emph{system of interest} whose behaviors we observe over time $t$.
We describe the set of behaviors as a \emph{stochastic process} $\Process$. We
require that the process is \emph{nonanticipatory}: There exists a natural
\emph{filtration}---i.e., an ordered (discrete or continuous) set of indices
$t\in\filtSet$ associated to time---such that the observed behaviors of the
system of interest only depend on past times. 

$\Process$'s observed behaviors are described by random variables $X$, 
indexed by $t$ and denoted by a capital letter. A particular
\emph{realization} $X_t = x \in \Xvals$ is denoted via lowercase letter.
We assume $X$ is defined on the measurable space $\left(\Xvals, \Borel^\Xvals,
\nu^\Xvals\right)$, where the domain $\Xvals$ of $X$'s values could be a
discrete alphabet or a continuous set. $\Xvals$ is endowed with a
reference measure $\nu^\Xvals$ over its Borel sets $\Borel^\Xvals$. 
In particular, $\Process$'s measure $\nu^\Xvals$ applies to 
time blocks: $\{ \Pr(X_{a< t \leq b}): a < b, a,b\in \filtSet \}$. 

The \emph{process} $\left( X_{t}\right)_{t\in\filtSet}$ is the main object 
of study. We associate the \emph{present} to a specific time $t\in\filtSet$ 
that we map to $0$, without loss of generality. We refer to a process' 
\emph{past} $X_{t\leq 0}$ and its \emph{future} $X_{t>0}$. In particular,
we allow $\filtSet=\mathbb{R}$ and infinite past or future sequences.

Beyond the typical setting in which an observation $x$ is a simple scalar, here
realizations $X_{t=0}=x$ can encode, for example, a vector $x=[ (m_i
)_{t\leq 0}]_{i=1...M}$ of $M$ measured time series
$(m_i)_{t\leq 0}$, each from a different sensor $m_i$, up to $t=0$. Often,
though, there are reasons to truncate such observation templates to a
fixed-length past. Take, for example, the case of exponentially decaying
membrane potentials where $x$ measures a neuron's electric activity\citep{Brodu07} or a
lattice of spins with decaying spatial correlations. After several decay
epochs, the templates can often be profitably truncated.

To emphasize, this contrasts with related works---e.g., Ref.
\onlinecite{klus2020}---in that we need not consider $X_{t}$ to be only the
presently-observed value nor, for that matter, $Y_{t}=X_{t+1}$ to be its
successor in a time series.  Rather, $X_{t}$ could be the full set of
observations up to time $t$. This leads to a substantial conceptual broadening,
the consequences of which are detailed in Refs. \onlinecite{Crut12a,Shal98a,Crut88a}.

\subsection{Predictions}
\label{subsec:Predict}

As just argued, we consider that random variable $X_t$ contains
information available from observing a system up to time $t$. At this
stage, we just have given data and we have made \emph{no} assumptions 
about its utility for prediction.
Consider another random variable $Y$ that describes system observations we 
wish to predict from $X$. A common example would be future sequences (possibly
truncated, as just noted) that occur at times $t>0$. We also assume $Y$ is
defined on a measurable space $\left(\Yvals,\Borel^{Y},\nu^{Y}\right)$. 

A \emph{prediction} then is the distribution of outcomes $Y$ given observed
system configuration $X=x$ at time $t$, denoted $\Pr\left(Y|X_t = x\right)$.
The same definition extends to nontemporal predictive settings by changing 
$t$ to the relevant quantity over which observations are collected;
e.g., indices for pixels in an image.

\subsection{Process Types}
\label{subsec:Types}

A common restriction on processes is that they are stationary---the same
realization $x_{t}$ at different times $t$ occurs with the same
probability. 

\begin{Def}[Stationarity]
\label{def:Stationarity}
A process $\left( X_{t}\right)_{t\in\filtSet}$ is \emph{stationary} if, at all
times $t' \neq t$, the same distribution of its values is observed:
$\Pr\left(X_{t'} = x\right) \equiv \Pr\left(X_{t}=x\right)$, for all $x \in
\Xvals$ and except possibly on a null-measure set.
\end{Def}

The following, in contrast with this common assumption, does not require
$\left( X_{t}\right)_{t\in\filtSet}$ to be stationary. In point of fact, such an
assumption rather begs the question. Part of discovering a process's structure
requires determining from data if such a property holds. However, we assume the
following from realization to realization.

\begin{Def}[Conditional~Stationarity]
\label{def:CondlStationarity}
A process $\left( X_{t}\right)_{t\in\filtSet}$ is \emph{conditionally
stationary} if, at all times $t' \neq t$, the same conditional distribution of
next values $Y$ is observed: $\Pr\left(Y|X_{t'} = x\right) \equiv
\Pr\left(Y|X_{t}=x\right)$, for all $x \in \Xvals$ and except possibly on a
null-measure set.
\end{Def}

That is, conditional stationarity allows marginals $\Pr\left(X_t\right)$ to
depend on time. But at \emph{any given time}, for each realization of the 
process where the same $x$ is observed, then the same conditional distribution 
is also observed. In a spatially extended context, the hypothesis could
also encompass realizations at different locations for which the same $X=x$ is
observed.

When multiple realizations of the same process are not available, we will
instead consider a limited form of \emph{ergodicity}: except for measure zero
sets, any one realization, measured over a long enough period, reveals the
process' full statistics. In that case, observations of the same $X=x$ can be
collected at multiple times to build a unique distribution
$\Pr\left(Y|X=x\right)$, now invariant by time-shifting. It may be that the
system undergoes structural changes: for example, a volcanic chamber slowly
evolving over the course of a few months or weather patterns evolving as the
general long-term climate changes. Then, we will only assume that the
$\Pr\left(Y|X=x\right)$ distribution is stable over a long enough time window
to allow its meaningful inference from data. It may slowly evolve over longer
time periods. 

If multiple realizations of the same process are available, this hypothesis may
not be needed. Then, only Def. \ref{def:CondlStationarity}'s conditional
stationarity is required for building the system states according to the method
we now introduce.

\subsection{Causal states}
\label{subsec:Causal-states}

\newcommand{\CS}{\CausalState}
\newcommand{\cs}{\causalstate}
\newcommand{\CSSet}{\CausalStateSet}

Finally, we come to the workhorse of computational mechanics that forms the
basis on which a process' structure is identified.

\begin{Def}[Predictive equivalence]
\label{def:PredEq}
Realizations $x$ and $\altx$ that lead to the same predictions (outcome
distributions) are then gathered into classes $\Eps(\cdot)$ defined by the
\emph{predictive equivalence relation} $\CausalEquivalence$: 
\begin{align}
\Eps(x) = \left\{ \altx \in \Xvals:
\Pr \left(Y|X= \altx \right) \equiv \Pr \left(Y|X=x\right)\right\}
  ~.
\label{eq:equivalence_relation}
\end{align}
\end{Def}

In other words, observing two realizations $x$ and $\altx$ means the process is
in the same effective state---same equivalence class---if they lead to the same
predictions:
\begin{align}
x ~\CausalEquivalence~ \altx
  \iff \Pr \left(Y|X=x\right) \equiv \Pr \left(Y|X=\altx \right)
  ~.
\end{align}

Speaking in terms of pure temporal processes, two observed pasts $x$ and
$\altx$ that belong to the same predictive class $\Eps(\cdot)$ are
operationally indistinguishable. Indeed, by definition of the conditional
distribution over futures $Y$, no new observation can help discriminate whether
the causal state arose from past $\altx$ or $x$. For all practical purposes,
these pasts are equivalent, having brought the process to the same condition
regarding future behaviors.

\begin{Def}[Causal states \citep{Crut88a}]
\label{def:CausalStates}
Since the classes $\{ \Eps(\cdot): x \in \Xvals\}$ induce the same
consequences---in particular, the same behavior distribution $\Pr\left(Y|X =
x\right) = \Pr\left(Y|\Eps(x)\right)$---they capture a process' internal
causal dependencies. They are a process' \emph{causal states} $\cs \in
\CausalStateSet$.
\end{Def}

Predictive equivalence can then be summarized as \emph{the same causes lead to
the same consequences}. Thanks to it, grouping a process' behaviors $\Xvals$
under the equivalence relation $\CausalEquivalence$ also gives the minimal
partition required for optimal predictions: further refining the classes of
pasts is useless, as just argued. While, at the same time, each class is
associated to a unique distribution of possible outcomes.

This setup encompasses both deterministic observations, where $Y = f(X)$ is
fixed and $\Pr(Y|X=x)$ becomes a Dirac distribution with support $f(x)$, as
well as stochastic observations. The source of stochasticity need not be
specified: fundamental laws of nature, lack of infinite precision in
measurements of a chaotic system, and so on.

Beyond identifying a process' internal structure, predictive equivalence is
important for practical purposes: it ensures that the partition induced by
$\CausalEquivalence$ on $\Xvals$ is stable through time. Hence, data observed
at different times $t_i$ may be exploited to estimate the causal states. In
practice, if the longterm dynamic changes, we assume predictive equivalence
holds over a short time window.

While causal states are rigorously defined as above, in empirical circumstances
one does not know all of a system's realizations and so one cannot extract its
causal states. Practically, we assume that the given data consists of a set of
$N$ observations $\left(x_{i},y_{i},t_{i}\right)_{i=1\ldots N}$. These data
encode, for each configuration $x_i$ at time $t_i$, what subsequent
configuration $y_i$ was then observed. The goal is to recover from such data
an approximate set of causal states that model the system evolution;
see, e.g., Refs. \onlinecite{Stre13a,Marz14f}.

\subsection{Causal-State Dynamic for Discrete Sequences}
\label{subsec:discrete_dynamics}

\newcommand{\SeqLenX}{L^X}
\newcommand{\SeqLenY}{L^Y}

For now, assume that the data $x \in \Xvals$ is a \emph{past}---a sequence
$x=\left(v_{-\SeqLenX<t\leq 0}\right)$ of discrete past values
$v_{t}\in\mathcal{V}$ at discrete times $t \leq 0$. This discrete setting helps
introduce several important concepts, anticipating generalizations that follow
shortly. Indices $t$ are now discrete observation times, ranging from
$\SeqLenX$ in the past up to the present at $t=0$. We allow
$\SeqLenX=\infty$ when calculating causal states analytically on known
systems. Similarly, we take $y\in \Yvals$ to be a \emph{future}---a sequence
$\left(v_{0<t\leq \SeqLenY}\right)$ of discrete future values $v\in\Yvals$
that may also be truncated at time $\SeqLenY$ in the future.

For discrete-value processes, $\mathcal{V}$ becomes an alphabet of symbols $v$
and $X$ and $Y$ both become semi-infinite (or truncated) sequences over
that alphabet. The transition from time $t_{0}$ to time $t_{1}$ is associated
with the observation of a new symbol $v\in\mathcal{V}$.

It should be noted that a surrogate space $\mathcal{V}$ can always be obtained
from continuous-value data if regularly sampled at times $t_{i}$, for
$t_{i+1}=t_{i}+\Delta t$ with a fixed $\Delta t$. In this case, pairs
$\left(x_{i},x_{i+1}\right)$ are equivalent to observing transition symbols
$v$. There are at most $\left|\mathcal{V}\right|$ such possible transitions
from the current causal state $\cs_{0} = \Eps\left(x_0\right)$ to which
$x_{0}=\left(v_{t\leq0}\right)$ belongs.

Conditional symbol-emission probabilities $\Pr\left(v\in\mathcal{V}|\cs\right)$
are defined (and can be empirically estimated) for each causal state $\cs$ and
for each emitted symbol $v\in\mathcal{V}$ \citep{Shal98a}. Since $\cs \in
\CausalStateSet$ encodes all past influence, by construction state-to-state
transitions do not depend on previous states. That is, causal states are
Markovian in this sense. The dynamic over causal states is then specified by a
set of symbol-labeled causal-state transition matrices $\{
T_{\cs,\cs^\prime}^{(v)}: ~\cs, \cs^\prime \in \CausalStateSet, x \in
\mathcal{V}\}$.

Diagrammatically, the causal states $\cs$ form the nodes of a directed graph
whose edges are labeled by elements of $v\in\mathcal{V}$ with associated
transition probabilities $\Pr\left(v|\cs\right)$. Moreover, the transitions are
\emph{unifilar}: the current state $\cs_0$ and next symbol $\nu_0$ uniquely
determine the next state $\cs_1 = f(\cs_0,\nu_0)$.

\subsection{\EMs}
\label{subsec:eMs}

\begin{figure}
\begin{centering}
\includegraphics[width=1\columnwidth]{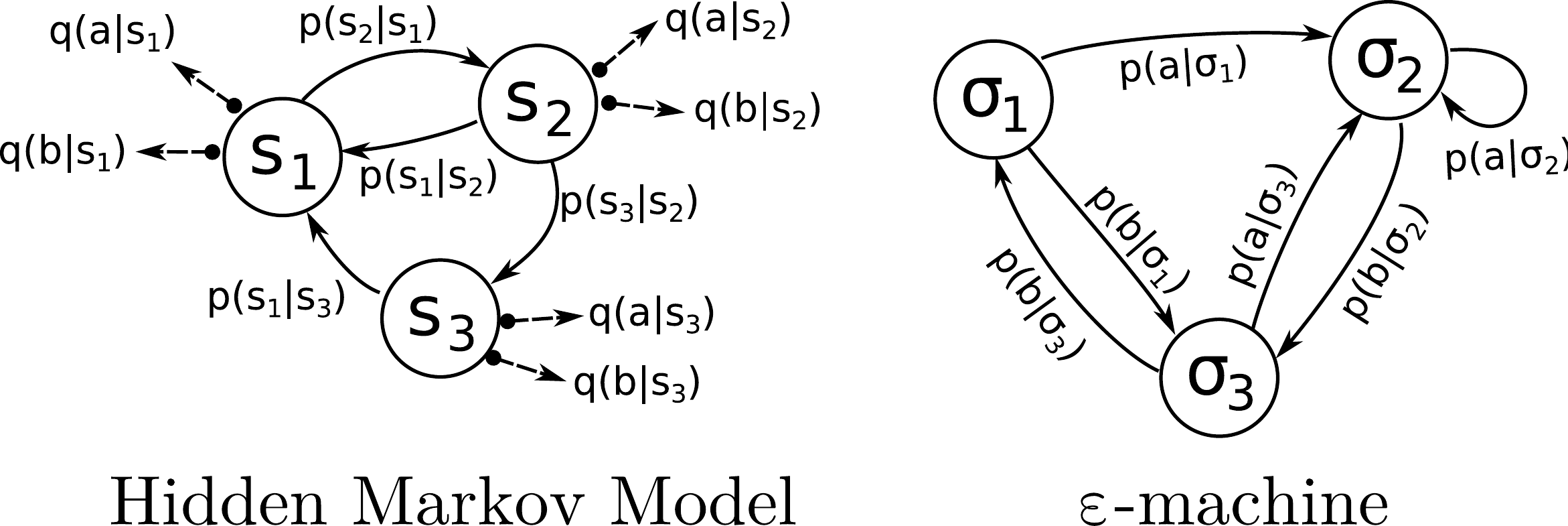}
\par\end{centering}
\caption{(Left) State-emitting Hidden Markov models: State transition
	probabilities $\Pr(s_i|s_j)$ are specified independently from the
	symbol-emission probabilities $q(a|s_i)$ and $q(b|s_j)$.
	(Right) \EMs: Symbols are emitted on transitions and the (causal) states
	capture dependencies.
	Unfortunately, for state-emitting HMMs the number of hidden states is a
	poor proxy for structural complexity and is often a meta-parameter with low
	interpretability. Since \eM is unique, so it directly represents a
	stochastic process' intrinsic properties, such as generated randomness
	(Shannon entropy rate) and structural complexity (memory).
  }
\label{fig:hmm-emachine}
\end{figure}

Taken altogether, the set of causal states and their transition dynamic define
the \emph{\eM}.\citep{Crut88a} Graphically, they specify the state-transition
diagram of an edge-emitting, unifilar \emph{hidden Markov model} (HMM). These
differ in several crucial ways from perhaps-more-familiar state-emitting HMMs.
\cite{Elli95a} For example, symbol emission in the latter depends on the state
and is independent of state-to-state transition probabilities; see Fig.
\ref{fig:hmm-emachine}.

The principle reason for using an \eM is that it is a process'
\emph{optimally-predictive} model.\citep{Shal98a} Specifically, an \eM is the
minimal, unifilar, and unique edge-emitting HMM that generates the process.

Notably, \eMs' Markov property is derived from the predictive equivalence
relation, thanks to the latter's conditioning on pasts. More generally, the
causal states are an intrinsic property of a process. They do \emph{not}
reflect a choice of representation, in contrast to how state-emitting HMMs are
often deployed; again see Fig. \ref{fig:hmm-emachine}. The same holds for state
transitions and symbol emissions, all of which are uniquely inferred from the
process.

Given that they make markedly fewer and less restrictive assumptions, it is not
surprising that reconstruction algorithms for estimating \eMs from data are
more demanding and costly to estimate
\citep{CSSR-UAI-2004,Brodu11,Goerg13,rupe2019disco} than performing a standard
expectation-maximum estimate for an hypothesized HMM.\cite{Elli94a} Most \eM
algorithms rely on a combination of clustering empirically-estimated sequence
probability distributions $\Pr\left(Y|X\right)$ together with splitting the
candidate causal states when required for consistency (unifilarity) of unique
emissions $\Pr(v\in\mathcal{V}|\cs \in \CausalStateSet)$.\cite{Shal02a,Shal04a}
To date, though, Bayesian inference for \eMs provides the most reliable
estimation and works well on small data sets.\cite{Stre13a}

We mentioned that \eMs are unifilar edge-emitting HMMs. Smaller
\emph{nonunifilar} HMMs can exist that are not predictive, but rather
\emph{generate} data with the same statistical properties as the given process.
However, one cost in using these smaller HMMs is the loss of determinacy for
which state a process is in, based on observations. The practical consequence
is that the processes generated by nonunifilar HMMs typically have an
uncountable infinity of causal states. This forces one to use probabilistic
\emph{mixed states}, which are distributions over the states of the generating
HMM. References \onlinecite{Jurg20b,Jurg20c,Jurg20d} develop the theory of mixed
states for nonunifilar, generative HMMs. For simplicity, the following focuses
on processes with finite ``predictive'' models---the \eMs. That said, Sec.
\ref{subsec:Mixed-states} below analyzes an inference experiment using a
process with an uncountable infinity of mixed states to probe the algorithm's
performance.

\subsection{Patterns Captured by \EMs}
\label{subsec:Patterns}

An \eM's hidden states and transitions have a definite meaning and allow for
proper definitions of process structure and organization---indicators of a
process' complexity.\citep{Crut12a} For example, we can calculate in
closed-form various information-theoretic measures to quantify the information
conveyed from the past to the future or that stored in the present.
\citep{Crut08b} In this way, \eMs give a very precise picture of a process'
information processing abilities.\citep{Crut08a,Riec18b}

More specifically, each causal state's surprisal can be used to build powerful
data filters.\citep{Hans95a,McTa04a,shalizi2006filters,Rupe17b,rupe2019disco}
The entropy of the causal states---the cost of coding them, the
\emph{statistical complexity}---can be used in entropy-complexity surveys of
whole process families to discriminate purely random from chaotic data.
\citep{Crut12a} Recent advances in signal processing \citep{Riec19a}
show how processes with arbitrarily complex causal structures can still exhibit
a flat power spectrum, since the  spectrum is the Fourier transform of only the
two-point autocorrelation function. This demonstrates the benefit of inferring
process structure using the full setup presented above---consider $X_{t}$
encompassing all information up to time $t$ and not restricting $X_{t}$ to a
present observation.\citep{Riec18b,Riec19a}

However, these benefits have been circumscribed. Many previous \eM inference
methods work with symbolic (discrete value) data in discrete time. However, in
practice often we monitor continuous physical processes at arbitrary sampling
rates and these measurements can take a continuum of data values within a range
$\mathcal{V}$. In these cases, estimation algorithms rely on clustering of
causal states by imposing arbitrary boundaries between discretized groups.
However, there may be a fractal set or continuum of causal states.
\citep{Marz17a} More recent approaches consider continuous time but keep
discrete events, such as renewal and semi-Markov processes.
\citep{Marz14c,Marz14e,Marz17b}

The method introduced below is, to our knowledge, the first that is able to
estimate causal states for essentially arbitrary data types and to represent
their dynamic in continuous time. This approach offers alternative algorithms
that provide a radically different set of assumptions and algorithmic
complexity than previous approaches. While it is also applicable to the
discrete case (see Sec. \ref{subsec:The-Even-Process}), it vastly expands
computational mechanics' applicability to process classes that were previously
inaccessible.

\section{Constructing Causal States using Reproducing Kernels}

The predictive-equivalence relation implicates conditional distributions with
expressing a process' structure. To work directly with conditional
distributions, this section recalls the main results concerning the geometric
view of probability distributions as points in a reproducing-kernel Hilbert
space. Following the method from Refs. \onlinecite{Smola07} and \onlinecite{Song09}, we
describe both unconditional and conditional distributions. Both are needed in
computational mechanics. Once they are  established in the RKHS setting, we
then describe the geometry of causal states.

\subsection{Distributions as Points in a Reproducing Kernel Hilbert Space}
\label{subsec:RKHS_intro}

Consider a function $k^{X}$ of two arguments, defined on
$\mathcal{X}\times\mathcal{X}\rightarrow\mathbb{R}$ (resp., $\mathbb{C}$).
Fixing one argument to $x\in\mathcal{X}$ and leaving the second free, we
consider $k^{X}(x,\cdot)$ as a function in the Hilbert space $\mathcal{H}^{X}$
from $\mathcal{X}$ to $\mathbb{R}$ (resp., $\mathbb{C}$). If $k^{X}$ is a
positive symmetric (resp., sesquilinear) definite function \citep{Aronszajn50},
then the \emph{reproducing property} holds: For any function
$f\in\mathcal{H}^{X}$ and for all $x\in\mathcal{X}$, we have $\left\langle
f,k(x,\cdot)\right\rangle _{\mathcal{H}^{X}}=f(x)$. Here, $\left\langle \cdot,
\cdot\right\rangle _{\mathcal{H}^{X}}$ is the inner product in
$\mathcal{H}^{X}$ or a \emph{completion} of $\mathcal{H}^{X}$; see
Ref. \onlinecite{Aronszajn50}. $\mathcal{H}^{X}$ is known as the
\emph{reproducing-kernel Hilbert space} (RKHS) associated with \emph{kernel}
$k^{X}$. 

Kernel functions $k^{X}$ are easy to construct and so have been defined for a
wide variety of data types, including vectors, strings, graphs, and so on. A
product of kernels corresponds to the direct product of the associated Hilbert
spaces \citep{Aronszajn50}. Thus, products maintain the reproducing property.
Due to this, it is possible to compose kernels when working with heterogeneous
data types.

Kernels are widely used in machine learning. A common use is to convert a given
linear algorithm, such as estimating \emph{support vector machines}, into
nonlinear algorithms.\citep{boser92svm} Indeed, when a linear algorithm can be
written entirely in terms of inner products, scalings, and sums of observations
$x$, then it is easy to replace the inner products in the original space
$\mathcal{X}$ by inner products in the Hilbert space $\mathcal{H}^{X}$. The
$k^{X}(x,\cdot)$ functions are then called \emph{feature maps} and
$\mathcal{H}^{X}$ the \emph{feature space}.

Returning to the original space $\mathcal{X}$, the algorithm now works with
kernel evaluations $k^{X}\left(x_{1},x_{2}\right)$ whenever an inner product
$\left\langle k^{X} \left(x_{1},\cdot\right), k^{X}\left(x_{2}, \cdot\right)
\right \rangle_{\mathcal{H}^{X}}$ is encountered. In this way, the linear
algorithm in $\mathcal{H}^{X}$ has been converted to a nonlinear algorithm in
the original space $\mathcal{X}$. Speaking simply, what was nonlinear in the
original space is linearized in the associated Hilbert space. This powerful
``kernel trick'' is at the root of the probability distribution mapping that we
now recall from Ref. \onlinecite{Smola07}.

\subsubsection{Unconditional distributions}
\label{subsec:Unconditional-distributions}

Consider a probability distribution $\Pr(X)$ of the random variable $X$.
Then, the average map $\alpha\in\mathcal{H}^{X}$ of the kernel evaluations
in the RKHS is given by $\alpha=E_{X}\left[k^{X}(x,\cdot)\right]$.
Note that for any $f\in\mathcal{H}^{X}$:
\begin{align*}
\left\langle \alpha,f\right\rangle
  & = \left\langle E_{X}\left[k^{X}(x,\cdot)\right],f\right\rangle \\
  & = E_{X}\left[\left\langle k^{X}(x,\cdot),f\right\rangle \right] \\
  & =E_{X}\left[f(x)\right]
  ~.
\end{align*}
An estimator for this average map can be computed simply as $\widehat{\alpha} =
\frac{1}{N} \sum_{i=1}^{N}k^{X}(x_{i},\cdot)$. This estimator is consistent
\citep{Smola07} and so converges to $\alpha$ in the limit of
$N\rightarrow\infty$.

Consider now the two-sample test problem: We are given two sets of samples
taken from a priori distinct random variables $A$ and $B$, both valued in
$\mathcal{X}$ but with possibly different distributions $\Pr(A) = P^{A}$ and
$\Pr(B) = P^{B}$. Do these distributions match: $P^{A}=P^{B}$? This scenario is
a classical statistics problem and many tests were designed to address it,
including the Chi-square and the Kolmogorov-Smirnov tests.

Using the RKHS setup and the average map, a new test \citep{Gretton12} is
simply to compute the distance between the average maps using the Hilbert space
norm: $\left\Vert \alpha^{A}-\alpha^{B}\right\Vert _{\mathcal{H}^{X}}$. Under
suitable mild conditions on the kernel, it can be shown \citep{Gretton12} that
$\left\Vert \alpha^{A}-\alpha^{B}\right\Vert _{\mathcal{H}^{X}}=0$ if, and only
if, $P^{A}=P^{B}$ up to a set of points with null measure in $\mathcal{X}$.

This \emph{maximum mean discrepancy} (MMD) test is consistent and accurate.
Moreover, confidence levels can be obtained through bootstrapping or other
techniques described in \citep{Gretton12}. In practice, for samples $\left\{
a_{i}\right\}_{i=1..N}$ and $\left\{ b_{j}\right\} _{j=1..M}$, $\left\Vert
\widehat{\alpha}^{A}-\widehat{\alpha}^{B}\right\Vert _{\mathcal{H}^{X}}$ can be
computed via:
\begin{align*}
\left\Vert
  \widehat{\alpha}^{A}-\widehat{\alpha}^{B}
  \right\Vert_{\mathcal{H}^{X}}^{2}
  & = \left\langle \widehat{\alpha}^{A}-\widehat{\alpha}^{B},
  \widehat{\alpha}^{A}-\widehat{\alpha}^{B}\right\rangle \\
  & = \left\langle \widehat{\alpha}^{A},\widehat{\alpha}^{A}\right\rangle
  + \left\langle \widehat{\alpha}^{B},\widehat{\alpha}^{B}\right\rangle
  - 2\left\langle \widehat{\alpha}^{A},\widehat{\alpha}^{B}\right\rangle 
  ~,
\end{align*}
where inner products between
$\widehat{\alpha}^{A}=\frac{1}{N}\sum_{i=1}^{N}k^{X}(x_{i},\cdot)$
and $\widehat{\alpha}^{B}=\frac{1}{M}\sum_{j=1}^{M}k^{X}(x_{j},\cdot)$
can easily be developed into sums of kernels evaluations $k^{X}(x_{i},x_{j})$.

This test makes it possible to compare two distributions \emph{without} density
estimation and directly from data. For all practical purposes, under mild
technical conditions,\citep{Sriperumbudur10} a distribution $\Pr(X)$ of random
variable $X$ can then be represented as a point in the RKHS $\mathcal{H}^{X}$,
consistently estimated by the mean mapping $\widehat{\alpha}$. The RKHS norm
then becomes a true distance between probability distributions.

\subsubsection{Conditional distributions}
\label{subsec:Conditional-distributions}

Consider random variables $X$ and $Y$ with the same notations as
above. The joint variable $\left(X,Y\right)$ leads to a direct product
Hilbert space $\mathcal{H}^{X}\otimes\mathcal{H}^{Y}$,\citep{Aronszajn50}
with the product kernel:
\begin{align*}
k^{X,Y}((x,y),\cdot)=k^{X}(x,\cdot)k^{Y}(y,\cdot)
  ~.
\end{align*}
For functions $f\in\mathcal{H}^{X}$ and $g\in\mathcal{H}^{Y}$, a covariance
operator $C_{YX}:\mathcal{H}^{X}\rightarrow\mathcal{H}^{Y}$ can be defined such
that:
\begin{align*}
\left\langle g,C_{YX}f\right\rangle _{\mathcal{H}^{Y}}=E\left[f(X)g(Y)\right]
  ~.
\end{align*}
Similarly, for $C_{XX}$ in the case $Y=X$.

Then, under strong conditions on the kernel,\citep{Song09,Song13} we can
relate the conditional mean map in $\mathcal{H}^{Y}$ to the conditioning point
in $\mathcal{H}^{X}$ using:
\begin{align*}
E_{Y}\left[k^{Y}(y,\cdot)|X=x\right]
  = C_{YX}C_{XX}^{-1} k^{X}\left(x,\cdot\right)
  ~.
\end{align*}
The strong conditions can be relaxed to allow the use of a wide
range of kernels by considering a regularized version:
\begin{align*}
E_{Y,\varepsilon} \left[k^{Y}(y,\cdot)|X=x\right]
  = C_{YX}\left(C_{XX}+\varepsilon I\right)^{-1} k^{X}\left(x,\cdot\right)
  ~.
\end{align*}
This is a consistent estimator of the unregularized version when computed
empirically from samples.\citep{Fukumizu13}

As for the unconditional case,
\begin{align*}
s_{Y|X=x} = E_{Y} \left[k^{Y}(y,\cdot)|X=x\right]
\end{align*}
can be seen as uniquely representing the distribution $\Pr(Y|X=x)$,
up to a null measure set. 

The set $S = \left\{ s_{Y|X=x}\right\}_{x \in \mathcal{X}} \subset
\mathcal{H}^{Y}$ traces out all possible conditional distributions
$\Pr(Y|X=x)$, for all $x\in\mathcal{X}$. It inherits the RKHS norm from
$\mathcal{H}^{Y}$: $\left\Vert s_{1}-s_{2}\right\Vert _{\mathcal{H}^{Y}}$ is
well-defined for any $s_{1}, s_{2} \in S$. Note that
$s_{Y|X=x}=\varsigma(x)$ can be also be interpreted as a injective function
$\varsigma:\mathcal{X} \rightarrow S$, selecting points in the RKHS
$\mathcal{H}^{Y}$ for each $x$.

The connection with the regression problem for estimating $\widehat{s}_{Y|X=x}$
from data,\citep{Grunewalder12,Song13} together with the representer theorem,
\citep{Scholkopf01} ensure that $\widehat{s}_{Y|X=x}$ lies in the data span:
\begin{align*}
\widehat{s}_{Y|X=x}=\sum_{i=1}^{N}\omega_{i}(x)k^{Y}(y_{i},\cdot)
\end{align*}
with $N$ the number of samples. For all practical purposes, then, dimension $N$
is sufficient when working with $S \subset \mathcal{H}^{Y}$. This is in
contrast to working with the infinite dimension of the underlying
$\mathcal{H}^{Y}$. The unconditional case shown in the previous section can be
understood when $\omega_{i}(x) = 1 / N$, in which case no dependency on $x$
remains.

A regularized and consistent estimator \citep{Fukumizu13,Song13,Song09}
is given by:
\begin{align*}
\omega(x)=\left(G^{X}+\varepsilon I\right)^{-1}K(x)
  ~,
\end{align*}
with $G_{ij}^{X} =k ^{X}\left(x_{i},x_{j}\right)$ the Gram matrix of the $X$
samples and $K(x)$ a column vector such that $K_{i}(x)=k^{X}(x,x_{i})$. Thanks
to Ref. \onlinecite{Grunewalder12}, $\varepsilon$ can be set by cross-validation.

In practice, it is more efficient to equivalently solve the linear system:
\begin{align}
\left(G^{X}+\varepsilon I\right)\omega(x)
  = K(x)
\label{eq:conditional_sample_weights-1}
\end{align}
to find the vector $\omega(x)$. Note that, for an invertible matrix and without
regularization, $\omega(x)$ is simply the vector with the only nonnull entry
$\omega_{i}=1$ for $x=x_{i}$. In practice, there may be duplicate entries
(e.g., for discrete data) or nearby values for which the regularization becomes
necessary.

In this way, employing a suitable kernel and regularization, a conditional
probability distribution $\Pr\left(Y|X=x\right)$ is represented in the RKHS
$\mathcal{H}^{Y}$ as a point:
\begin{align*}
\widehat{s}_{Y|X=x}=\sum_{i=1}^{N}\omega_{i}(x)k^{Y}(y_{i},\cdot)
  ~,
\end{align*}
with a reasonably easy way to estimate the coefficients $\omega(x)$ from data.

In this light, the full $\Omega$ matrix obtained via:
\begin{align*}
\left(G^{X}+\varepsilon I\right)\Omega=G^{X}
\end{align*}
can also be seen as a way to ``spread'' the influence of the
$\left(x_{i},y_{i}\right)$ observations to nearby $x_{i}$ values, so that all
duplicate $x_{i}$ effectively belong to the same estimated conditional
distribution.

It is also possible to convert a conditional embedding back into a true density
over $\mathcal{X}$ using RKHS preimage techniques;
\citep{honeine09rkhs_preimage} the most advanced one to date is
Ref. \onlinecite{schuster20kdo}.

\subsection{RKHS Causal States}
\label{subsec:states_measures_rkhs}

Section \ref{subsec:Causal-states} defined a process' causal states $\cs \in
\CausalStateSet$ via the predictive equivalence relation:
\begin{align*}
\Eps(x) =
  \left\{ w\in\mathcal{X}:\,\Pr\left(Y|X=w\right)
  = \Pr\left(Y|X=x\right)\right\} 
  ~.
\end{align*}
This is exactly the preimage $E=\varsigma^{-1} \left(s_{Y|x}\right) \subset
\mathcal{X}$ of the unique point $s_{Y|E}\in\mathcal{H}^{Y}$, with $s_{Y|E} =
s_{Y|w}$ for all $w \in E = \Eps\left(x\right)$. Therefore, we can refer
equivalently to one or the other concept: we refer to the set in $\mathcal{X}$
by the equivalence class and the point in the RKHS as $s\in\mathcal{H}^{Y}$.

In short, the set $S = \left\{s_{Y|X=x}\right\}_{x\in\Xvals} \subset
\mathcal{H}^{Y}$, for $x \in \mathcal{X}$, is the set\CausalStateSet of causal
states. Note that $S$ is a subset of all possible conditional
probability-distribution mappings in $\mathcal{H}^{Y}$. It consists of only the
mappings that actually correspond to some $x$ in the domain of interest. We now
drop referring to $S$ and refer only to causal states $\cs \in \CausalStateSet$
with random variable $\CS$.

Let $\mathcal{B}$ be the Borel sets over $\CausalStateSet$. For any set
$\beta\in\mathcal{B}$, its preimage in $\mathcal{X}$ is defined by:
\begin{align*}
\varsigma^{-1}(\beta)
  = \bigcup_{\cs\in\beta}\left\{ x\in\varsigma^{-1}(\cs)\right\} 
  ~.
\end{align*}
Recall that, by definition, the causal states form a partition of
$\mathcal{X}$. Hence, each $x$ belongs to a unique preimage
$\varsigma^{-1}(\cs)$, when the union is taken over $\cs \in \beta$ in the
preceding definition.

Recall that $\nu^{X}$ is the reference measure on $\mathcal{X}$, in terms of
which probability distribution $\Pr(X)$ is defined. A natural, push-forward
measure $\mu$ is defined on $\left(\CausalStateSet,\mathcal{B}\right)$ by:
\begin{align*}
\mu\left(\beta\in\mathcal{B}\right)=\nu(\varsigma^{-1}(\beta))
  ~. 
\end{align*}

If $\Pr(X)$ admits a probability density function $p =
d\Pr(X) / d\nu^{X}$---the \emph{Radon--Nikodym derivative} of $\Pr(x)$
with respect to $\nu^{X}$---then we can similarly push-forward the density on
$\mathcal{X}$ to define a density of causal states:
\begin{align*}
q\left(\cs\right)=\int_{x\in\varsigma^{-1}(\cs)}p\left(x\right)d\nu^{X}
  ~.
\end{align*}
The distribution $Q$ of states over $\CausalStateSet$ is then defined by:
\begin{align*}
Q\left(\beta\in\mathcal{B}\right)
  & = \int_{\cs\in\beta}q(\cs)d\mu_{\cs} \\
  & = \int_{\cs\in\beta}
  \left(
  \int_{x \in \varsigma^{-1}(\cs)} p \left(x\right) d\nu^{X}
  \right) d\mu_{\cs}
  ~.
\end{align*}

Note that the measure $\mu$ is defined on (and restricted to) $\CausalStateSet$
and that no equivalent of the Lebesgue measure exists for the whole
infinite-dimensional space $\mathcal{H}^{Y}$.

The net consequence is that causal states are to be viewed simultaneously as:
\begin{itemize}
\item Sets of points in $\mathcal{X}$, using the predictive equivalence
	relation $\sim_\Eps$, with equivalence classes forming a partition of
	$\mathcal{X}$. Though this is the original definition, presented in
	Section \ref{subsec:Causal-states}, it is still very useful in the RKHS
	setup to define the push-forward measure.
\item Conditional probability distributions of possible outcomes $Y$ given
	observed system configurations $X$. This view is used in clustering
	algorithms \citep{Brodu11,Goerg13} to identify causal states. These
	algorithms directly map to the RKHS setting by using the MMD
	test instead of other statistical tests for clustering conditional
	distributions.
\item Points in the RKHS $\mathcal{H}^{Y}$: More specifically, points in the
	subset $\CausalStateSet = \left\{ \cs_{Y|X=x}\right\}_{x\in\Xvals}$. The
	subset $\CausalStateSet$ (and only $\CausalStateSet$, not the rest of
	$\mathcal{H}^{Y}$) is endowed with a push-forward measure $\mu$. Thus, we
	can properly define probability distributions and densities of causal
	states in this Hilbert space setting.
\end{itemize}
Compared to previous works, this third alternate view offers several
advantages:
\begin{itemize}
\item (Nearly-)arbitrary data types can be handled though the use of
	reproducing kernels. We are no longer limited to discrete alphabets
	$\mathcal{V}$ of symbolic values. Adequate kernels exist for many data
	types and can even be composed to work directly with heterogeneous data
	types.
\item The distance $\left\Vert \cdot\right\Vert _{\mathcal{H}^{Y}}$ can
	serve as the basis for clustering similar states together. Thanks to the
	MMD test introduced in Section \ref{subsec:Unconditional-distributions},
	algorithms need not rely on estimating the conditional densities
	$\Pr(Y|X=x)$ for clustering states. 
\end{itemize}

\section{Dynamics of RKHS causal states}
\label{sec:Dynamics-of-causal-states}

The preceding constructed causal states $\cs \in \CausalStateSet$ statically.
Computational mechanics aims at modeling a process' behaviors, though, not just
describing its constituents. A process' dynamic induces trajectories over
$\CausalStateSet$. The next sections describe them and then define a new class
of predictive models---the \keMs---and explain how to use them for computing
statistics of the given data, how to simulate new ``data'', and how to predict
a process.

\subsection{Causality and Continuity}
\label{subsec:Causality-and-continuity}

Consider a series of causal states $\ldots \cs_{t-1} \cs_{t} \cs_{t+1}
\ldots$. Distinct pasts $x \in \cs_{0}$ and $w \in \cs_{0}$ both contain the
same information concerning the process dynamics, since they induce the same
distribution of futures $\Pr\left(Y|\cs_{0}\right)$. However, $\cs_0$ does not
contain the information about which past is responsible for arriving at it:
whether $x$ or $w$ or any other past led to the same state $\cs_{0}$.

So, what kind of causality is entailed by causal states? Causal states capture
the same global information needed to describe how the future is impacted by
the present, and they consist of all possible ways that information was
organized in the past. Unifilarity can now be understood as a change of state
$\delta \cs$ uniquely determined by the information gain. In contrast,
the equivalent of nonunifilarity would be that despite having all useful
knowledge about the past (previous state) and having all possible information
about the change of system configuration (either $\delta x$ or $\delta w$,
and so on), the model states are still not defined unequivocally. This is not the case with the definition of the causal states, but would be with other
generative models, for which only a distribution of states can be inferred from data.\citep{Marz17a}

The continuity of causal state trajectories can be understood from an information-theoretic perspective. The relative information $d_{KL} \left(x_{t+dt}||x_{t}\right)$ (Kullback-Leibler divergence) between $x_{t}$ and its evolution after an infinitesimal time $t+dt$ is simply the change of information we have on $x$. Assuming that information comes only at finite velocity, then $d_{KL}\left(x_{t+dt}||x_{t}\right)\rightarrow0$ as $dt\rightarrow0$. However, it is known that $d_{KL}\left(x_{t+dt}||x_{t}\right)\rightarrow0$ implies $\left\Vert x_{t+dt}-x_{t}\right\Vert _{\mathcal{H}^{X}}\rightarrow0$.\citep{Sriperumbudur10} Using
Sec. \ref{subsec:Conditional-distributions}'s construction of conditional
distributions, we find that $\left\Vert \cs_{t+dt}-\cs_{t}\right\Vert
_{\mathcal{H}^{Y}}\rightarrow0$ as $dt\rightarrow0$ with continuous kernels and
positive definite operators. Hence, causal-state trajectories are continuous. 

In practice, assumptions leading to continuous trajectories need not be
respected:
\begin{itemize}
\item For practical reasons, it is rarely possible to work with infinite
	series. Truncating the past and the future can be necessary. In these
	cases, there is information not fully contained in $\widehat{x}$---the
	truncated estimate of $x$. Truncation can be justified when the causal
	influence of past system configurations onto the present decays
	sufficiently fast with time; then we ignore old configurations with
	negligible impact. (And, similarly for the future.) This amounts to saying
	there are no long-range correlations, that no useful information for
	prediction is lost when ignoring these past configurations. Hence, that
	$\Pr(Y|X)$ is unchanged by the truncation. This hypothesis mail fail, of
	course, with the consequence that truncating the past or future may
	introduce jumps in the estimated trajectories of the causal states in
	$\CausalStateSet$---a jump induced by the sudden loss or gain of
	information.
\item When using continuous kernel functions, the associated RKHS norm is
	also continuous in its arguments: $\left\Vert \cs_{Y|X=x_{1}} -
	\cs_{Y|X=x_{2}}\right\Vert _{\mathcal{H}^{Y}}\rightarrow0$ as
	$x_{1}\rightarrow x_{2}$. Hence, continuity in data trajectories also
	implies continuity in causal-state trajectories. However, when data are
	truly discrete in nature, the situation of Sec.
	\ref{subsec:discrete_dynamics} is recovered. An alternate view is that, at
	a fundamental level, information comes in small packets: these
	are the symbols induced by the data changes in this discrete scenario.
\item It may also be that measurements are performed at a finite time scale
	$\tau$. Then, the information gained between two consecutive measurements
	can be arbitrarily large, but still appear instantaneously at the scale at
	which data is measured. This leads to apparent discontinuities.
\end{itemize}
Here, we discuss only the  elements needed to address causal states in
continuous time with nearly arbitrary data; i.e., for which a reproducing
kernel exists. A full treatment of discontinuities is beyond the present
scope and best left for the future.

\subsection{Continuous Causal-State Trajectories}
\label{subsec:Continuous_Trajectories}

From here on, assume that trajectories of causal states $\cs \in
\CausalStateSet$ are continuous. Recall, though, that $\CausalStateSet \subset
\mathcal{H}^{Y}$---a metric space. This guarantees the existence of a reference
Wiener measure $\omega$ on the space of trajectories defined over a time
interval $[0,t<T]$ (possibly $T=\infty$) with a canonical Wiener process $W$.
With $\cs_{t}$ being a state-continuous Markov process, we posit that its
actual trajectories evolve under an It{\^o} diffusion---a stochastic differential
equation (SDE) of the form:
\begin{align}
d\cs_{t}  =  a\left(\cs_{t}\right)dt+b\left(\cs_{t}\right)dW_{t}
  ~,
\label{eq:ito_diffusion}
\end{align}
where $a\left(\cs_{t}\right)$ is a (deterministic) \emph{drift} and
$b\left(\cs_{t}\right)$ a \emph{diffusion}. Both depend on the current state
$\cs_{t}$, hence this It{\^o} diffusion is inhomogeneous in state space. However,
the coefficients are stable in time $a(\cs_{t},t)=a(\cs_{t})$ and
$b(\cs_{t},t)=b(\cs_{t})$ if the (possibly limited) ergodicity assumption is
made in addition to the conditional stationarity assumption of Sec.
\ref{subsec:Causal-states}. And so, the diffusion is homogeneous in time. This
SDE is the equivalent of the \eM's causal-state-to-state transition
probabilities introduced in the discrete case (Sec.
\ref{subsec:discrete_dynamics}). The evolution equation encodes, for each
causal state $\cs_{t}$, how the process evolves to nearby states at $t+dt$.

The causal state behavior arises from integrating over time:
\begin{align*}
\cs_{t} = \cs_{0} +
  \int_{0}^{t} a \left(\cs_{\tau}\right) d\tau +
  \int_{0}^{t} b \left(\cs_{\tau}\right) dW_{\tau}
\end{align*}
for any state $\cs_{0}$ being the trajectory's initial condition. As for
the evolution of the causal-state distribution, consider a unit probability
mass initially concentrated on $\delta\left(\cs_{0}\right)$. Then the states
$\cs_{t}$ reached at time $t$ define a probability distribution
$\Pr\left(\beta\in\mathcal{B}|\cs_{0},t\right)$ on $\CausalStateSet$, with
associated density $p\left(\cs_{t}|\cs_{0}\right)$. (Recall that the latter is
defined as $p=d \Pr(\cdot)/d\mu$ with $\mu$ the push-forward measure from
$\Xvals$; see Sec. \ref{subsec:states_measures_rkhs}.) This distribution
encodes the state-to-state transition probabilities at any time $t$,
parallel to iterating an \eM in the discrete case.

The evolution of probability densities $p$ other than $\delta \left( \cs_{0}
\right)$ is governed by a Fokker-Planck equation. The infinitesimal generator
$\Gamma$ of the process $\cs_t$ is:
\begin{align*}
\Gamma f\left(\cs_{0}\right) = \lim_{t\rightarrow0}
  \frac{\mathbb{E} \left[f\left(\cs_{t}\right)|\cs_{0},t\right]
  - f\left(\cs_{0}\right)}{t}
  ~,
\end{align*}
where $\Gamma$ is an operator applied on a suitable class of functions
$f$.\footnote{For It{\^o} diffusions, these $f$ are compactly-supported and twice
differentiable.} For a state distribution $Q$, with associated density
$q=dQ/d\mu$, the Fokker-Planck equation corresponding to the It{\^o} diffusion can
be written in terms of the generator's adjoint $\Gamma^{*}$: 
\begin{align}
\frac{\partial q}{\partial t}\left(\cs,t\right)
  = \Gamma^{*} q \left(\cs,t\right)
  ~.
\label{eq:Fokker-Planck_using_Generator}
\end{align}

Restricting to the data span in $\CausalStateSet$ (see Sec.
\ref{subsec:Conditional-distributions}), using the representer theorem, and the
samples as a pseudo-basis,\citep{Scholkopf01} the operator $\Gamma$ can be
represented using a vector equation in a construction similar to classical
euclidean $\mathbb{R}^{N}$ state spaces, see Ref.~\onlinecite[p.~103]{Jacobs10}. 
There are, however, practical difficulties associated with estimating 
coefficients $a$ and $b$ (using tangent spaces) at each observed state 
$\cs_{t}$ in RKHS.

As an alternative, we represent the generator in a suitable functional basis,
on which $q$ is also represented as a set of coefficients. Then the state
distribution evolves directly under:
\begin{align}
q(\cs,t) = e^{\left(t-t_{0}\right) \Gamma^{*}} q \left(\cs,t_{0}\right)
  ~.
\label{eq:fokker-planck-solution}
\end{align}

The evolution operator $E\left(t\right)=e^{\left(t-t_{0}\right)\Gamma^{*}}$ can
also be inferred from data for standard SDE over $\mathbb{R}^{N}$, as detailed
in \citep{berry2015nonparametric}. The next section summarizes the method and
adapts it to the RKHS setting.

Every It{\^o} diffusion also admits a limit distribution $L\left(\cs \in
\CausalStateSet \right)$ as $t \rightarrow \infty$, with associated density
$\ell=dL/d\mu$. By definition of the limit distribution, $\ell$ is the
eigenvector of $E$ associated to eigenvalue $1$: $E\left(t\right)[\ell]=\ell$.

The limit distribution can also be useful to compute how ``complex'' or
``uncommon'' a state is, using its self-information $h(\cs) = -\log \ell(\cs)$.
This is the equivalent of how the statistical complexity is defined in the
discrete case \citep{Shal98a}. And, it can be used for similar purposes (e.g.,
building filters \citep{Hans95a,shalizi2006filters,Rupe17b,rupe2019disco}).
That said, this differential entropy should be interpreted with caution.

\subsection{Diffusion Maps and the Intrinsic Geometry of Causal States}
\label{subsec:Diffusion-maps}

The representer theorem \citep{Scholkopf01} allows us to develop causal-state
estimates $\widehat{\cs}$ in the span of the data kernel functions
$k^{Y}\left(y_{i},\cdot\right)$, used as a pseudo-basis; see Sec.
\ref{subsec:Unconditional-distributions}. Since the span dimension 
grows with the number of samples, estimation algorithms are impractical.
However, the causal states are an intrinsic property of the system, independent
of how they are coordinatized. And so, when working with data acquired from
physical processes, $\CausalStateSet$ will appear as the dominant structure.
The question then becomes how to work with it.

The following assumes $\CausalStateSet$ is of small finite dimension $M$
compared to the number of observations $N$.\footnote{It may be that
$\CausalStateSet$'s dimension is actually infinite, so that its estimate grows
with the number of samples. This can be detected using the spectral method
presented in the main text.} Residing on top of the dominant structure, the
statistical inaccuracies in the MMD test appear as much smaller contributions.
The following, thus, introduces the tools needed to work directly on $\CSSet$,
whether for visualizing the causal states using reduced coordinates or for
representing evolution operators on $\CSSet$ instead of using
Eq.~(\ref{eq:ito_diffusion}) on $\mathcal{H}^{Y}$.

To do this, we exploit the methodology introduced for diffusion maps.
\citep{Coifman06} These maps are especially relevant when data lie on a curved
manifold, where the embedding's high-dimensional distance between two data
points does not reflect the manifold's geometry. In contrast, diffusion maps,
and their variable-bandwidth cousin \citep{berry16vbdm}, easily recover the
Laplace-Beltrami operator for functions defined on the manifold. Assuming
$\CausalStateSet$ is a smooth Riemannian manifold, then, the diffusion
coordinates cleanly recover $\CausalStateSet$'s intrinsic geometry (a static
property) independent of the observed local sampling density (a dynamic
property, linked to trajectories as in Sec.
\ref{sec:Dynamics-of-causal-states}).\citep{Coifman06} 

The original diffusion maps method artificially builds a proximity measure for
data points using a localizing kernel; i.e., one that takes nonnegligible
values only for near neighbors. Path lengths are computed from these
neighborhood relations. Two points are deemed ``close'' if there is a large
number of short paths connecting them, and ``far'' if there are only a few
paths, or paths with long distances, connecting them. See Ref.
\onlinecite{Coifman06} for
details, which also shows that the diffusion distance is a genuine metric.

That said, in the present context, there already is a notion of proximity
between causal states $\cs \in \CSSet$. Indeed, reusing notation from Sec.
\ref{subsec:Conditional-distributions}, the state estimates in RKHS are of the
form:
\begin{align*}
\widehat{\cs}_{Y|X=x} = \sum_{\alpha=1}^{N} \omega_{\alpha}(x)
  k^{Y}(y_{\alpha},\cdot)
  ~.
\end{align*}
And so, a Gram matrix $G^{\CSSet}$ of their inner products can be defined easily:
\begin{align}
G_{ij}^{\CSSet} & = \left\langle \widehat{\cs}_{Y|X=x_{i}}, \widehat{\cs}_{Y|X=x_{j}}
  \right\rangle \nonumber \\
G_{ij}^{\CSSet}i & =\left\langle \sum_{\alpha=1}^{N} \omega_{\alpha}(x_{i})
  k^{Y}(y_{\alpha},\cdot), \sum_{\beta=1}^{N} \omega_{\beta}(x_{j})
  k^{Y}(y_{\beta},\cdot)\right\rangle \nonumber \\
G_{ij}^{\CSSet} & = \sum_{\alpha=1}^{N} \sum_{\beta=1}^{N}
  \omega_{\alpha}(x_{i}) \omega_{\beta}(x_{j})
  k^{Y}(y_{\alpha},y_{\beta})
   ~,
\label{eq:GS_linear_sum_kernels}
\end{align}
where $k^{Y}(y_{\alpha},y_{\beta})=G_{\alpha\beta}^{Y}$ is the Gram matrix of
the $Y$ observations. The determination of the $\omega$ coefficients for each
$x$ relies on the $G^{X}$ gram matrix, as detailed in Sec.
\ref{subsec:Unconditional-distributions}.

It turns out that diffusion maps can be built from kernels with exponential
decay.\citep{berry16local_kernels} The original fixed-bandwidth diffusion maps
\citep{Coifman06} also use exponential kernels for building the proximity
relation. Such kernels are reproducing and they are also characteristic,
fulfilling the assumptions needed for representing conditional distributions.
\citep{Fukumizu13}. Hence, when using the exponential kernel, the RKHS Gram
matrix is also exactly a similarity matrix that can be used for building a
diffusion map. (This was already made explicit in Ref. \onlinecite{Coifman05}). Moreover,
Eq.  (\ref{eq:GS_linear_sum_kernels}) explicitly represents $G_{ij}^{\CSSet}$
as a weighted sum of exponentially decaying kernels and, hence, is itself
exponentially decaying. Thus, $G^{\CSSet}$ can be directly used as a similarity
matrix to reconstruct $\CSSet$'s geometry via diffusion maps.

Notice, though, that here data is already scaled by the reproducing kernel. So,
for example, using a Gaussian kernel:
\begin{align*}
k^{X}\left(x_{1},x_{2}\right) = \exp
  \left(-\left\Vert x_{1}-x_{2} \right\Vert_{\mathcal{X}}^{2}/\xi^{2}\right)
  ~,
\end{align*}
$\xi$ specifies the scale at which differences in the $\Xvals$ norm are
relevant. Similarly for $k^{Y}$ and $\mathcal{Y}$. Since that scale $\xi$ can
be set by cross-validation,\citep{Grunewalder12} we exploit this fact shortly
in Sec. \ref{subsec:Lorenz-Attractor}'s experiments when an objective measure
is provided; e.g., prediction accuracy.

In practice, a large range of $\xi$ can produce good results and an automated
method for selecting $\xi$ has been proposed.\citep{coifman08autotune}
Varying the analyzing scale to coarse-grain the manifold $\CSSet$ is also
possible. Using the method from Ref. \onlinecite{coifman05multiscale}, this is similar,
in a way, to wavelet analysis.

Once the data scale is properly set and the similarity matrix built,
the diffusion map algorithm can be parametrized to cleanly recover
$\CSSet$'s Riemannian geometry, doing so independently from how
densely sampled $\CSSet$ is. This is explained
in Ref.~\onlinecite[Fig.4]{Coifman06} and it is exactly what is needed need here:
separate $\CSSet$'s static causal-state geometric structure
from the dynamical properties (trajectories) that induce the
density on $\CSSet$. This is achieved by normalizing the similarity
matrix $G^{\CSSet}$ to remove the influence of the sampling density,
then applying a spectral decomposition to the nonsymmetric normalized
matrix; see Ref. \onlinecite{Coifman06}.

The result is a representation of the form:
\begin{align}
\cs_{i} \equiv \left(
  \lambda_{1}\psi_{i,1}, \ldots, \lambda_{N} \psi_{i,N}
  \right)
\label{eq:dmap}
  ~,
\end{align}
where each $\psi_{\alpha}$, $\alpha=1\ldots N$, is a right eigenvector
and each $\lambda_{\alpha}$ is the associated eigenvalue. Note that
coefficients $\cs_{i,j}=\lambda_{j}\psi_{i,j}$ are also weights for
the conjugate left eigenvectors $\Phi_{j=1\ldots N}$, which are themselves
functions of the RKHS. (Hence, they are represented in practice by the $N$
values they take at each sample.)

The first eigenvalue is $1$ and it is associated with constant right
eigenvector coordinates.\citep{Coifman06} The conjugate left eigenvector
coefficients yield an estimate of the limit density $\ell
\left(\widehat{\cs}_{i}\right)$, with respect to the reference measure $\mu$ in
our case. Hence, $\Phi_{1}$ and $\psi_{1}$ can be normalized so that
$\sum_{j}\Phi_{1,j}=1$ and $\psi_{1,j}=1$ for all $j$. With these conventions,
$\lambda_{1}\psi_{1,j}=1$ is constant and can be ignored, all the while
respecting the bi-orthogonality $\left\langle \psi_{1},\Phi_{1}\right\rangle
=1$. The other eigenvalues are all positive $1>\lambda_{\alpha}>0$ and we can
choose the indexing $\alpha$ so they are sorted by decreasing order. 

When $\CSSet$ is a low-dimensional manifold, as assumed here, a spectral gap
should be observed. Then, it is sufficient to retain only the $M\ll N$ first
components. Otherwise, $M$ can be set so that the residual distance
$\sum_{\alpha>M} \lambda_{\alpha}^{2} \left( \psi_{i,\alpha} - \psi_{j,\alpha}
\right)$ remains below some threshold $\theta$. Since the diffusion distance is
a true metric in the embedding space $\mathcal{H}^{Y}$, $\theta$ can also be
set below a prescribed significance level for the MMD test (Sec.
\ref{subsec:Unconditional-distributions}), if so desired. The residual
components are then statistically irrelevant. 

Taking stock, the preceding established that:
\begin{enumerate}
      \setlength{\topsep}{0pt}
      \setlength{\itemsep}{0pt}
      \setlength{\parsep}{0pt}
\item The causal-state manifold $\CSSet$ can be represented in terms of
	a functional basis $\left\{ \Phi\right\} _{m=1\dots M}$ in $\mathcal{H}^{Y}$ of
	reduced dimension $M$. This is in contrast to using the full data span
	$\left\{ k_{i}^{Y}\left(y_{i},\cdot\right)\right\}_{\mathcal{H}^{Y}}$ of
	size $N$. The remaining components are irrelevant.
\item The functional basis $\left\{ \Phi\right\} _{m=1\dots M}$ can be defined in
	such a way that the induced representation of $\CSSet$ does not depend on
	the density at which various regions of $\CSSet$ are sampled. This, cleanly
	recovers $\CSSet$'s geometry.
\item Each causal state $\cs$ is represented as a set of coefficients in that
	basis.
\end{enumerate}
Taken altogether, the RKHS causal states and diffusion-map equations of motion
define a new structural model class---the \emph{\keMs}. The constituents inherit
the desirable properties of optimality, minimality, and uniqueness of \eMs
generally and provide a representation for determining quantitative properties
and deriving analytical results. That said, establishing these requires careful
investigation that we leave to the future.

\section{Deploying \KeMs}
\label{subsec:DeploukeMs}

As with \eMs generally, \keMs can be used in a number of ways. The following
describes how to compute statistics and how to make predictions.

\subsection{Computing Functions of Given Data}
\label{subsec:functions_of_data}

To recover the expected values of functions of data a functional can be defined on the reduced basis. Recall that, thanks to the
reproducing property:
\begin{align*}
\left\langle \cs,f \right\rangle
  = \mathbb{E}_{P\left(Y|\cs\right)}\left[f(x\in \cs)\right]
   ~,
\end{align*}
for any $f \in \mathcal{H}^{X}$. Such functions $f$ are represented in practice
by the values they take on each of the observed $N$ samples, with the
reproducing property only achieved for $N\rightarrow\infty$ samples (and
otherwise approximated). One such is $f_{\tau}$ that, for each observed past
$x$ associates the entry of a future time series $y$ matching time $\tau$. This
function can be fit from observed data. It is easy to generalize to
spatially-extended or network systems or to any function of the
$\left(x_{i},y_{i}\right)$ data pairs. However, $\left\langle
\cs,f\right\rangle $ can be expressed equally well in the reduced basis
$\left\{ \Phi\right\} _{m=1..M}$. Then, $f_{\tau}$ is simply projected
$f_{m}=$$\left\langle f_{\tau},\Phi_{m}\right\rangle$ onto each eigenvector. 

This leads to an initial way to make predictions:
\begin{enumerate}
      \setlength{\topsep}{0pt}
      \setlength{\itemsep}{0pt}
      \setlength{\parsep}{0pt}
\item For each data sample, represent the function $f$ by the value it takes
	for that sample. For example, for vector time series of dimension $D$, $x$
	is a past series ending at present time $t_{0}$ and $y$ a future series.
	$f_{\tau}$ is then the $D$ values observed at time $t_{0}+\tau$ in the data
	set, for each sample, yielding a $N\times D$ matrix.
\item Project the function $f$ to the reduced basis by computing
	$f_{m}=\left\langle f_{\tau},\Phi_{m}\right\rangle$ for each left
	eigenvector $m\leq M$. This yields a $M\times D$ matrix representation
	$\widehat{f}$.
\item Compute $\widehat{f}\left[\cs_{i}\right]$ for a state $\cs_{i}$, itself
	represented as a set of $M$ coefficients in the reduced basis
	(Eq.~(\ref{eq:dmap})). This yields a $D$-dimensional vector in the original
	data space $\mathcal{X}$ in this example. This can be compared to the
	actual value from $y_{i}$ at time $\tau$.
\end{enumerate}

\subsection{Representing New Data Values}
\label{subsec:representing-new-data}

A model is useful for prediction if its states can be computed on newly
acquired data $x^{\mathrm{new}}$, for which future values $y$ are not
available. In the case of kernel methods and diffusion maps, Nystr{\"o}m extension
\citep{drineas05nystrom} is a well-established method that yields a reasonable
state estimate $\widehat{\cs}^{\mathrm{new}}$ if $x^{\mathrm{new}}$ lies within
a dense region of data. That said, it is known to be inaccurate in sparsely
sampled regions.

Given the Fokker-Planck evolution equation solutions
(Eq.~(\ref{eq:fokker-planck-solution})) and the evolution operator estimation
methods described shortly, we may estimate a distribution
$\widehat{q}^{\mathrm{new}}\left(\cs\right)$ over $\CSSet$, encoding the
probability that the causal state associated to $x^{\mathrm{new}}$ is at
location $\cs \in \CSSet$. Then, a single approximation
$\widehat{\cs}^{\mathrm{new}}=\mathbb{E}\left[\widehat{q}^{\mathrm{new}}
\left(\cs\right)\right]$ could be obtained, if desired. We can also allow the
distribution to degenerate to the Dirac $\delta$ distribution and yield
effectively a single estimate. This could occur, for example, when the
evolution is applied to one of the original samples $x^{\mathrm{new}}=x_{i}$
used for estimating the model.

To estimate a distribution $\widehat{q}^{\mathrm{new}}\left(\cs\right)$, we
employ the \emph{kernel moment matching},\citep{song08kernel_moment_matching}
adapted to our context. The similarity of the new data observation
$x^{\mathrm{new}}$ to each reference data $x_{i=1\ldots N}$ is computed using
kernel evaluations $K\left(x^{\mathrm{new}}\right) = \left\{
k^{X}\left(x^{\mathrm{new}},x_{i}\right)\right\} _{i=1\ldots N}$. Applying Eq.
(\ref{eq:conditional_sample_weights-1}) to the new vector
$K \left( x^{\mathrm{new}} \right)$:
\begin{align*}
\omega^{\mathrm{new}}
  = \text{argmin}_\omega
  \left|\left(G^{X}+\varepsilon I\right)\omega
  - K\left(x^{\mathrm{new}}\right)\right|^{2}
  ~,
\end{align*}
subject to $\omega_{i}^{\mathrm{new}} \geq 0$ for $i = 1, \ldots, N$.

Compared to Eq. (\ref{eq:conditional_sample_weights-1}) this adds a positivity constraint, similar to kernel moment matching \citep{song08kernel_moment_matching}. This also implies $\omega_{i}^{\mathrm{new}}\leq\left(1+\xi\right)/\left(1+\varepsilon\right)$,
where $\xi$ is the tolerance of the argmin solver. Proof: We know $R=G^{X}+\varepsilon I$ has $1+\varepsilon$ on the diagonal. Both $G^{X}$ and $K\left(x^{\mathrm{new}}\right)$ have positive entries. Then, $\left(1+\varepsilon\right)\omega_{i}+\sum_{j\neq i}R_{ij}\omega_{j}=k^{X}\left(x^{\mathrm{new}},x_{i}\right)\pm e$, where $e<\xi$ is the error of the argmin solver. However, $\sum_{j\neq i}R_{ij}\omega_{j}\geq0$, since $G_{ij}^{X}>0$ and $\omega_{j}\geq0$ by constraint. $k^{X}\left(x^{\mathrm{new}},x_{i}\right)\leq1$ by construction. Hence, $\left(1+\varepsilon\right)\omega_{i}\leq1+\xi$ and $\omega_{i}\leq\left(1+\xi\right)/\left(1+\varepsilon\right)$.

$\omega^{\mathrm{new}}$ is thus the closest solution to $\left(G^{X}+\varepsilon
I\right)^{-1}K\left(x^{\mathrm{new}}\right)$, so that the estimated state:
\begin{align*}
\widehat{\cs}^{\mathrm{new}}
  = \sum_{i=1}^{N} \omega_{i}^{\mathrm{new}} k^{Y} \left(y_{i},\cdot\right)
\end{align*}
remains in the convex set in $\mathcal{H}^{Y}$ of the data
$k^{Y}(y_{i},\cdot)$, up to a good approximation $\left(1+\xi\right) /
\left(1+\varepsilon\right)\approx 1$. Currently, lacking a formal justification
for convexity, we found that better results are obtained with it than with
Nystr{\"o}m extensions -- these use possibly negative
$\omega_{i}^{\mathrm{new}}$ values due to the unbounded matrix inverse,
yielding estimates that can wander arbitrarily (depending on $\varepsilon$) far
away in $\mathcal{H}^{Y}$ --. Normalizing, we get a probability distribution in
the form:
\begin{align*}
\widehat{q}^{\mathrm{new}} \left(\cs_{i}\right)
  = \frac{ \omega_{i}^{\mathrm{new}} }{ \sum_{j}\omega_{j}^{\mathrm{new}}}
  ~,
\end{align*}
where, as usual in RKHS settings, the function $\widehat{q}^{\mathrm{new}}$
is expressed as the values it takes on every reference data sample,
$\cs_{i=1\ldots N}$ in this case.

Compared to kernel moment matching \citep{song08kernel_moment_matching}, we
used the kernels $k^{Y}\left(y_{i},\cdot\right)$ as the basis for expressing
the density estimation, with coefficients that encode the similarity of the new
sample to each reference sample in $\Xvals$. An alternative would be to adapt
the method from \citep{schuster20kdo} and express $\widehat{q}^{\mathrm{new}}$
on reference samples drawn from the limit distribution $\ell$ over $\CSSet$; see
Sec. \ref{subsec:Continuous_Trajectories}.

When data is projected on a reduced basis $\left\{ \Phi\right\} _{m=1..M}$,
the distribution $\widehat{q}^{\mathrm{new}}$ can be applied to the reference
state coordinates $\cs_{i,m}=\lambda_{m}\psi_{i,m}$, so that an estimated state
$\widehat{\cs}^{\mathrm{new}} = \mathbb{E}\left[ \widehat{q}^{\mathrm{new}}
\left(\cs\right) \right]$ can be computed with:
\begin{align*}
\widehat{\cs}_{m}^{\mathrm{new}}
  = \left(\lambda_{m} \sum_{i}
  \widehat{q}_{i}^{\mathrm{new}}\psi_{i,m}\right)_{m=1\ldots M}
   ~.
\end{align*}

\subsection{Prediction with the Evolution Operator}
\label{subsec:Making-predictions}

Section \ref{subsec:functions_of_data}'s method allows predicting any arbitrary
future time $\tau$, provided that the future is sufficiently well represented
in the variable $y \in \Yvals$. In practice, this means that the future
look ahead $\SeqLenY$ needs to be larger than $\tau$ and that sufficient data is
captured to ensure a good reproducing capability for the kernel $k^{Y}$.
However, for some systems, the autocorrelation decreases exponentially fast and
there is no practical way to collect enough data for good predictions at large
$\tau$.

An alternative employs the Fokker-Planck equation to evolve state distributions 
over time. This, in turn, yields an estimate:
\begin{align*}
\mathbb{E}_{Q \left(\cs,t|t_{0}\right)} 
  \left[\mathbb{E}_{\Pr\left(Y|\cs\right)} \left[f(x\in \cs) \right] \right]
  ~.
\end{align*}
$Q$ here is the state distribution reached at time $t>t_{0}$, whose density is given by Eq.~(\ref{eq:fokker-planck-solution}). This method exploits $\CSSet$'s full structure, its Markovian properties, and the generator described in Sec. \ref{subsec:Continuous_Trajectories}.

This allows reaching longer times $t>\tau$, while using look aheads $\SeqLenY$
(Sec. \ref{subsec:discrete_dynamics}) that match the natural decrease of
autocorrelation. If the variable $y \in \Yvals$ captures sufficient information
about the system's immediate future for a given $x \in \Xvals$, then the causal
structure is consistently propagated to longer times by exploiting the
causal-state dynamics given by the evolution operator
$E\left(t\right)=e^{\left(t-t_{0}\right)\Gamma^{*}}$.

Thanks to expressing $\CSSet$ in the basis $\left\{ \Phi\right\} _{m=1\dots M}$, it
is possible to explicitly represent the coefficients $a\left(\cs_{t}\right)$
and $b\left(\cs_{t}\right)$ of Eq.~(\ref{eq:ito_diffusion}) in a traditional
vector representation, with established SDE coefficient estimation methods
\citep{Friedrich11,Jacobs10}. However, recent results suggest that working
directly with the evolution operator is more reliable
\citep{berry2015nonparametric,alexander20KAF,schuster20kdo} for similar models
working directly in data space---instead of, as here, causal-state space. 

Assuming states are estimated from observations acquired at regularly sampled
intervals, so that $\cs_{i+1}$ and $\cs_{i}$ are separated by time increment
$\Delta t$, then, in the functional basis $\left\{ \Phi\right\} _{m=1\dots M}$, the
coefficients $\psi_{i+1,m}$ are related to the coefficients $\psi_{i,m}$ by the
action of the evolution operator $E\left(\Delta t\right)$ on the state $s_{i}$.
Hence, this time-shifting operator from $t_{0}$ to time $t=t_{0}+\Delta t$ can
be estimated with:
\begin{align}
\widehat{E}\left(\Delta t\right) \propto \psi_{2:N}^{T} \Phi_{1:N-1}
\label{eq:evolution_operator_estimate}
  ~,
\end{align}
where $\psi_{2:N}$ is the set of $M$ right eigenvectors, restricted to times
$t\ge t_{0}+\Delta t$ and $\Phi_{1:N-1}$ are the corresponding $M$ left
eigenvectors, restricted to times $t\leq t_{0}+\left(N-1\right)\Delta t$.

Normalization can be performed a posteriori: $\widehat{E}\left(\Delta
t\right)\left[\widehat{\cs}\right]$ should have the constant $1$ as the first
coefficient (Sec. \ref{subsec:Diffusion-maps}). So, it is straightforward to
divide $\widehat{E}\left(\Delta t\right)\left[\widehat{\cs}\right]$ by its first
coefficient for normalization. The estimator $\widehat{E}\left(\Delta t\right)$
is efficiently represented by a $M\times M$ matrix. For a similar operator
working in data space $\mathcal{X}$, instead in $\CSSet$, the estimator is
consistent in the limit of $N\rightarrow\infty$, with an error growing as
$O\left(\sqrt{\Delta t/N}\right)$.\citep{berry2015nonparametric}

Predictions for values at future times $n\Delta t$, obtained by operator
exponentiation:
\begin{align*}
E\left(n\Delta t\right)
  & = e^{n\Delta t\Gamma^{*}} \\
  & = \left(e^{\Delta t\Gamma^{*}}\right)^{n}
  ~,
\end{align*}
are simply estimated with their matrix counterpart $\widehat{E}\left(n\Delta
t\right) = \widehat{E}\left(\Delta t\right)^{n}$. Thanks to the
bi-orthogonality of the left and right eigenvectors, this last expression can
be computed efficiently with:
\begin{align*}
\widehat{E}\left(n\Delta t\right)\propto\psi_{n:N}^{T}\Phi_{1:N-n}
\end{align*}
and a posteriori normalization. Though this estimator is statistically
consistent in the limit of $N\rightarrow\infty$, in practice, when $n\geq N$
the method clearly fails. This is counterbalanced in some cases by a
convergence towards the limit distribution in $n\ll N$ steps, so the problem
does not appear. This is the case for experiments in Sec.
\ref{subsec:Lorenz-Attractor} using a chaotic flow. Yet, the general case is a
topic for future research.

To synopsize, prediction is achieved with the following steps:
\begin{enumerate}
      \setlength{\topsep}{0pt}
      \setlength{\itemsep}{0pt}
      \setlength{\parsep}{0pt}
\item Represent a function $f$, defined on $\mathcal{X}$, by the values
	it takes for each observed data pair $\left(x_{i},y_{i}\right)$, with the
	method described in Sec. \ref{subsec:functions_of_data}. This gives an
	estimate $\widehat{f}$.
\item Build the evolution operator $\widehat{E}\left(\Delta t\right)$ as
	described above or powers of it $\widehat{E}\left(n\Delta t\right)$ for
	predictions $n$ steps in the future.
\item Compute the function $\widehat{E}\left(n\Delta
t\right)\left[\widehat{f}\right]$. This amounts to a matrix multiplication in
	the reduced basis representation, together with an a posteriori
	normalization.
\item When a new data sample $x^{\mathrm{new}}$ becomes available at the
	present time $t_{0}$, estimate a distribution $Q$ over the training samples
	that best represents $x^{\mathrm{new}}$. $Q$ is expressed by its density
	$\widehat{q}^{new}$ in the reduced basis $\left\{ \Phi\right\} _{m=1..M}$
	as detailed in see Section \ref{subsec:representing-new-data}.
\item Apply the evolved function
	$\widehat{E} \left( n\Delta t\right) \left[\widehat{f}\right]$ to
	$\widehat{q}^{new}$ to obtain the expected value
	$\mathbb{E}_{Q\left(\cs,t_{0}+n\Delta
	t\right)}\left[\mathbb{E}_{P\left(Y|\cs\right)}\left[f(x\in
	\cs)\right]\right]$ that $f$ takes at future time $t_{0}+n\Delta t$. 
\end{enumerate}

Section \ref{subsec:Lorenz-Attractor} applies this to a concrete example.

\section{Validation and Examples}
\label{sec:Validation-Examples}

The following illustrates reconstructing \keMs from data in three complementary
cases: (i) an infinite Markov-order binary process generated by a two-state
hidden Markov model, (ii) a binary process generated by an HMM with an
uncountable infinity of causal states, and (iii) thermally-driven
continuous-state deterministic-chaotic flows. In each case, the hidden causal
structure is discovered assuming only that the processes are conditionally
stationary.

\subsection{Infinite-range Correlation: The Even Process}
\label{subsec:The-Even-Process}

\begin{figure}
\includegraphics[width=1\columnwidth]{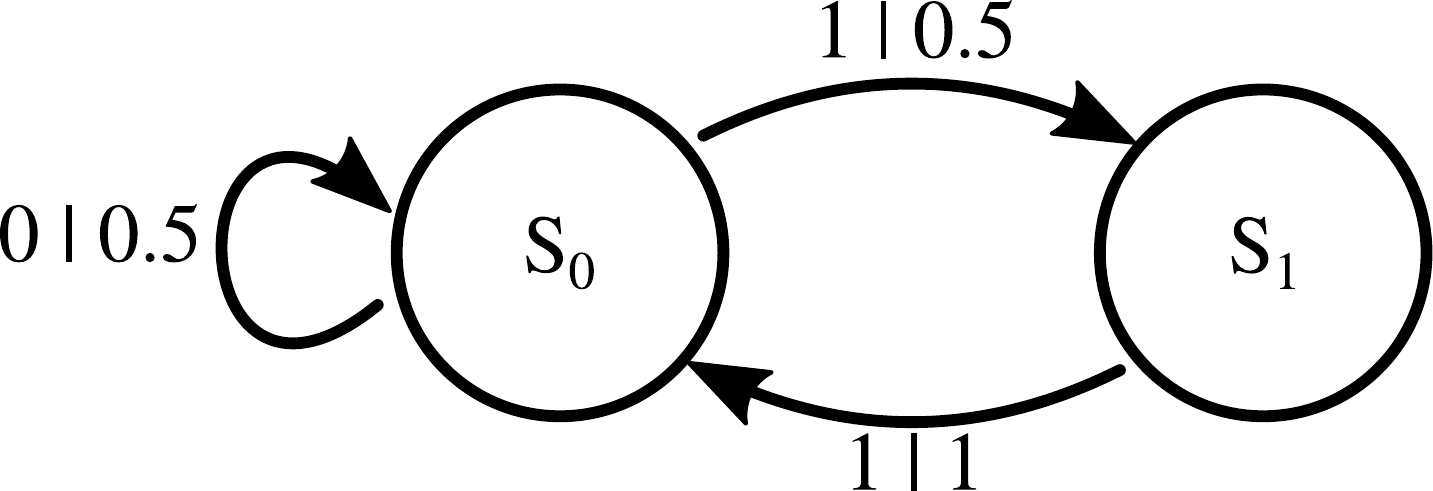}
\caption{Even Process state-transition diagram: An HMM that generates a binary
	process over outputs $v \in \{0,1\}$. Transitions are labeled with the
	symbol, followed by the probability to take this transition. The Even 
	Process has infinite Markov order---emitted	$1$s occur in even blocks (of 
	arbitrary length) bounded by $0$s. The
	process is stationary when starting with state distribution $\Pr(\CS =
	\cs_0, \CS = \cs_1) = (2/3, 1/3)$. This HMM is an \eM.
	}
\label{fig:Even-Process}
\end{figure}

The \emph{Even Process} is generated by a two-state, unifilar, edge-emitting
Markov model that emits discrete data values $v\in\{0,1\}$. Figure
\ref{fig:Even-Process} displays the \eM HMM state-transition diagram---states
and transition probabilities.

Realizations $x=\left(v_{t}\right)_{-\SeqLenX<t\leq 0}$ and
$y=\left(v_{t}\right)_{0 < t \leq \SeqLenY}$ consist of sequences in which blocks of even
number of $1$s are bounded by any number of $0$s; e.g.,
$01101111000011001111110\ldots$. An infinite-past look ahead $\SeqLenX$ is
required to correctly predict these realizations. Indeed, truncation generates
ambiguities when only $1$s are observed.

For example, with $\SeqLenX=4$ the observed series $\mathbf{1111}$ could be part of a
larger group $\ldots01\mathbf{1111}$, in which case the next symbol is
necessarily a $1$, or larger groups of the form $\ldots11\mathbf{1111}$ or
$\ldots00\mathbf{1111}$ or $\ldots10\mathbf{1111}$, in which case the next symbol
is either $0$ or $1$ with probability $1/2$. However, with a limited look ahead
of $\SeqLenX=4$, a prediction has no way to encode that the next symbol is
necessarily a $1$ in the first case. One implication is that there does not
exist any finite Markov chain that generates the process.

Despite its simplicity---only two internal states---and compared to processes
generated by HMMs with finite Markov order, the Even Process is a helpfully
challenging benchmark for testing the reconstruction capabilities of \keM
estimation on discrete data.

Reconstruction is performed using product kernels: 
\begin{align*}
k^{X}\left(x,\cdot\right)
  & = \prod_{i=0}^{\SeqLenX-1} k^{V}
  \left(x_{-i},\cdot\right)^{\gamma\frac{i}{\SeqLenX-1}} ~\text{and} \\
k^{Y}\left(y,\cdot\right)
  & = \prod_{i=1}^{\SeqLenY} k^{V}
  \left(y_{i},\cdot\right)^{\gamma\frac{i-1}{\SeqLenY-1}}
  ~,
\end{align*}
with a decay parameter $\gamma$ setting the relative influence of
the most distant past (future) symbol $x_{-\SeqLenX+1}$
($y_{\SeqLenY}$). We use the exponential kernel:
\begin{align*}
k^{V}\left(a,b\right) = e^{-\left(a-b\right)^{2}/2 \xi^{2}}
  ~,
\end{align*}
where $a,b\in\left\{ 0,1\right\}$ are the symbols in the series. 

\begin{figure}
\includegraphics[width=1\columnwidth]{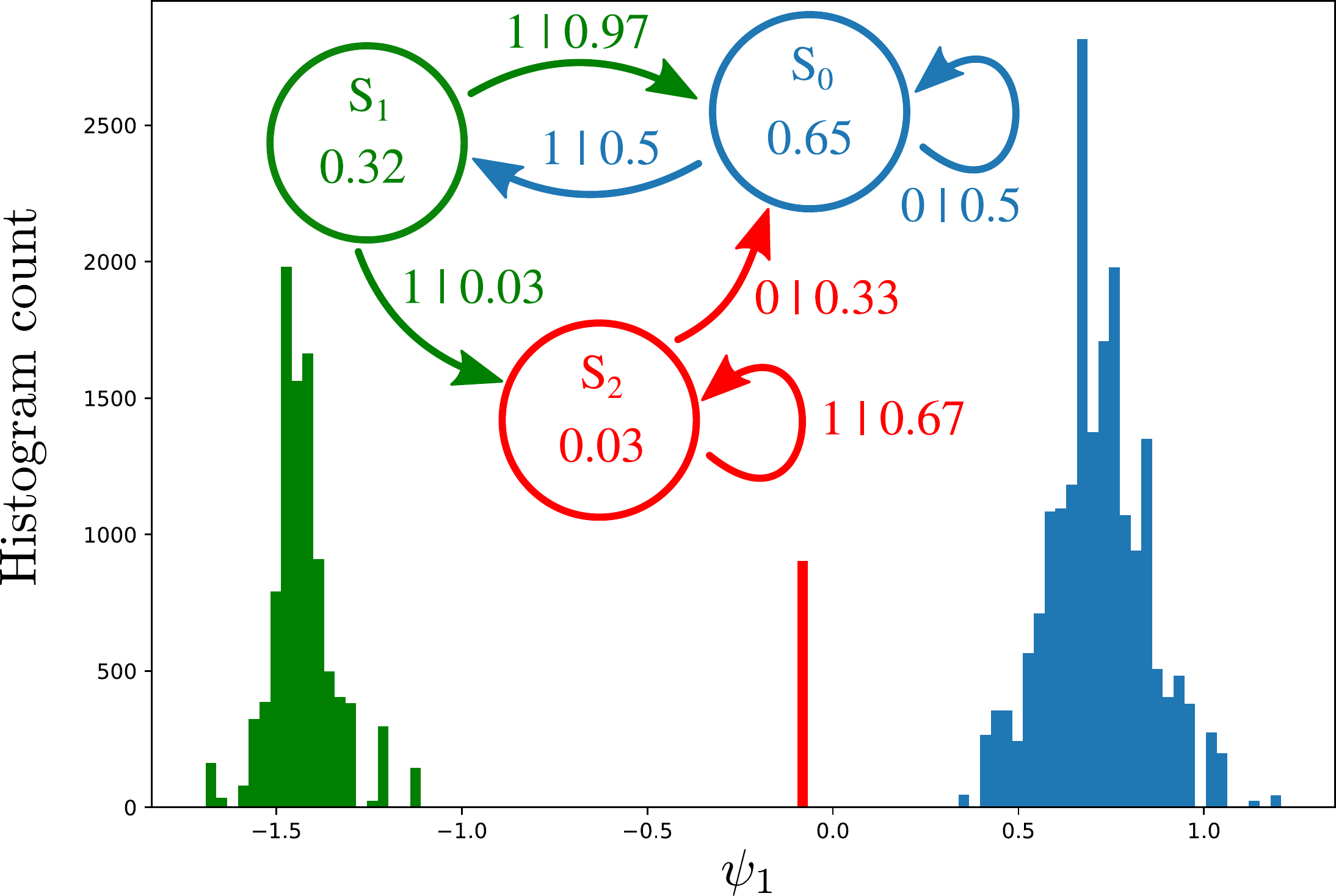}
\caption{Even Process: Reconstructed-state coordinates $\psi_{1}$ on the first
	reduced-basis eigenvector $\Phi_{1}$, together with a graphical
	representation of the transitions inferred between the colored clusters.
	}
\label{fig:EvenHistogram}
\end{figure}

Figure~\ref{fig:EvenHistogram} presents the results for $\SeqLenX=10$ and
$\SeqLenY=5$ for a typical run with $N=30,000$ sample $\left(x,y\right)$ pairs,
with a decay $\gamma=0.01$ and a bandwidth $\xi=1$.

The eigenvalue spectrum of the reduced basis decreases rapidly: $\lambda_{0}=1$
(as expected), $\lambda_{1}\approx10^{-2}$, and all other eigenvalues
$\lambda_{j \ge 2} < 10^{-4}$. We therefore project the causal states on the
only relevant eigenvector $\Phi_{1}$ and build the histogram shown in
Fig.~\ref{fig:EvenHistogram}. Colors match labels automatically found by a
clustering algorithm.\footnote{Here DBSCAN with threshold 0.1 was used.
However, any reasonable clustering algorithm will work given that clusters are
well separated.}

Figure~\ref{fig:EvenHistogram} summarizes the cluster probabilities and their
transitions.\footnote{The probabilities sum exactly to $1$. We show all the
transitions found.} They match the original Even Process together with a
transient causal state. By inspection, one sees this state represents the set
of sequences of all $1$s mentioned above---the source of ambiguity. Its
probability, for $\SeqLenX=10$, is that of jumping $5$ consecutive times from
state $\cs_{0}$ to state $\cs_{1}$ in the generating Even Process. Hence,
$1/2^{5} \approx 0.03$, which is the value we observe. From that transient
state, the ambiguity cannot be resolved so transitions follow the $(1/3,2/3)$
proportions of the symbols in the series. Note that unifilarity is broken,
since there are two paths for the symbol $v=1$ starting from state $\cs_{1}$,
also reflecting the ambiguity induced by the finite truncation.

\subsection{Infinite state complexity: An uncountable causal-state process}
\label{subsec:Mixed-states}

The Even Process is a case where ambiguity arises from incomplete knowledge, due
to finite-range truncation. However, even for discrete finite data alphabets
$\mathcal{V}$, there are process whose causal states are irreducibly
infinite. This occurs generically for processes generated by nonunifilar HMMs.
In this case, knowledge of the observed data is insufficient to determine the
generative model's internal state. Only distributions over those states---the
\emph{mixed states} \citep{Marz17a}---are predictive.

In the limit $\SeqLenX\rightarrow\infty$, the causal states then correspond to unique mixed-state distributions \citep{Crut08b} and there can be infinitely
many (countable or uncountable) causal states, even for a simple finite
generative HMMs. This arises from nonunifilarity since the same observed symbol
allows transitions to two distinct internal states. In contrast, the
predictive-equivalence determines that the same information (newly observed
symbol), starting from the same current state (equivalence class $\Eps(X)$ of
pasts $X \in \Xvals$), induces the same consequences (possible futures $Y$,
with a fixed distribution $\Pr\left(Y|\Eps(X)\right)$). Thus, nonunifilar
models can be more compact generators. This can be beneficial, but it comes at
the cost of being markedly-worse predictors than unifilar HMMs.

\begin{figure}
\includegraphics[width=1\columnwidth]{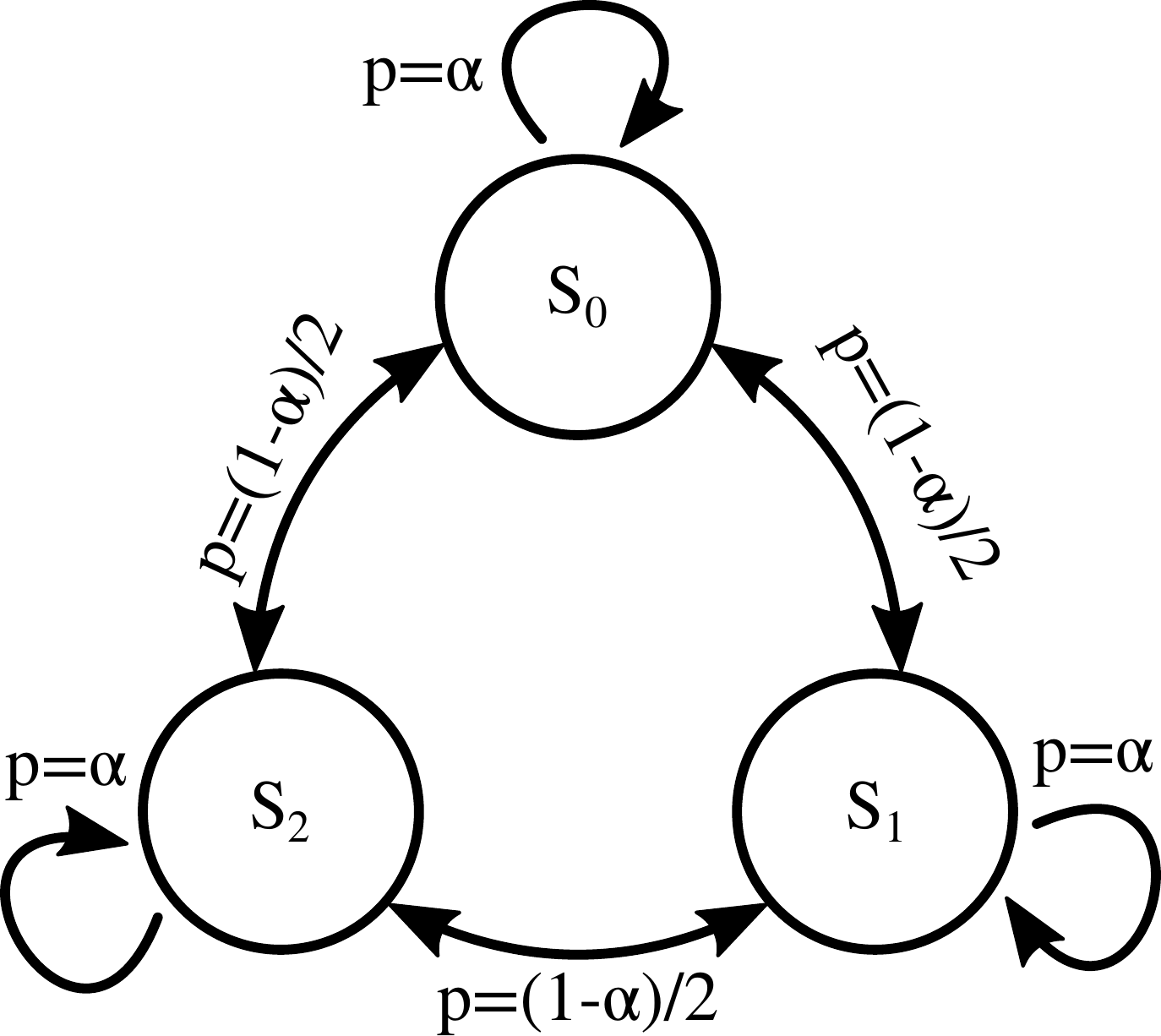}
\caption{Nonunifilar generative HMM mess3: Only transition probabilities
	are depicted; emitted symbols are described in the main text.
	}
\label{fig:mess3}
\end{figure}

Consider the nonunifilar \emph{mess3} HMM introduced in Ref.  \onlinecite{Marz17a}, represented graphically in Fig.~\ref{fig:mess3}.

Mess3 uses $3$ symbols $\mathcal{V}=\left\{ 0,1,2\right\}$ and consists of $3$
generative states $s_{0}$, $s_{1}$, and $s_{2}$. The state-transition diagram
is symmetric and each state has the same transition structure. Consider the
state $s_{i}$ for $i=0,1,2$ and modulo arithmetic so that $i-1=2$ when $i=0$
and $i+1=0$ when $i=2$. Then, the transitions are:
\begin{enumerate}
      \setlength{\topsep}{0pt}
      \setlength{\itemsep}{0pt}
      \setlength{\parsep}{0pt}
\item From state $s_{i}$ to itself, for a total probability $p=\alpha=ay+2bx$,
	symbols are emitted as $\Pr(v=i)=ay$ and $\Pr(v=i-1) = \Pr(v=i+1) = bx$.
\item From state $s_{i}$ to state $s_{i+1}$, for a total probability
	$p=(1-\alpha)/2$, symbols are emitted as $\Pr(v=i)=ax$, $\Pr(v=i-1)=bx$,
	and $\Pr(v=i+1)=by$.
\item From state $s_{i}$ to state $s_{i-1}$, for a total probability
	$p=(1-\alpha)/2$, symbols are emitted as $\Pr(v=i)=ax$, $\Pr(v=i-1)=by$,
	and $\Pr(v=i+1)=bx$.
\end{enumerate}
In this, $a=0.6$, $b=(1-a)/2=0.2$, $x=0.15$, $y=1-2x=0.7$.

The generated process is known to give uncountably-infinite causal states.
Their ensemble $\CSSet$ forms a Sierpinski gasket in the mixed-state simplex
\citep{Marz17a}. The probability to observe states at subdivisions
decreases exponentially, though. With limited samples, only a few
self-similar subdivisions of the main gasket triangle can be observed.

Reconstructing $\CSSet$ is performed using $\SeqLenX=15$ and $\SeqLenY=1$. The same
product exponential kernel is used as above, with a decay $\gamma=0.01$ and a
bandwidth $\xi=0.1$. $N=25,000$ sample $\left(x,y\right)$ pairs are
sufficient to reconstruct the first self-similar subdivisions.

\begin{figure}
\includegraphics[width=1\columnwidth]{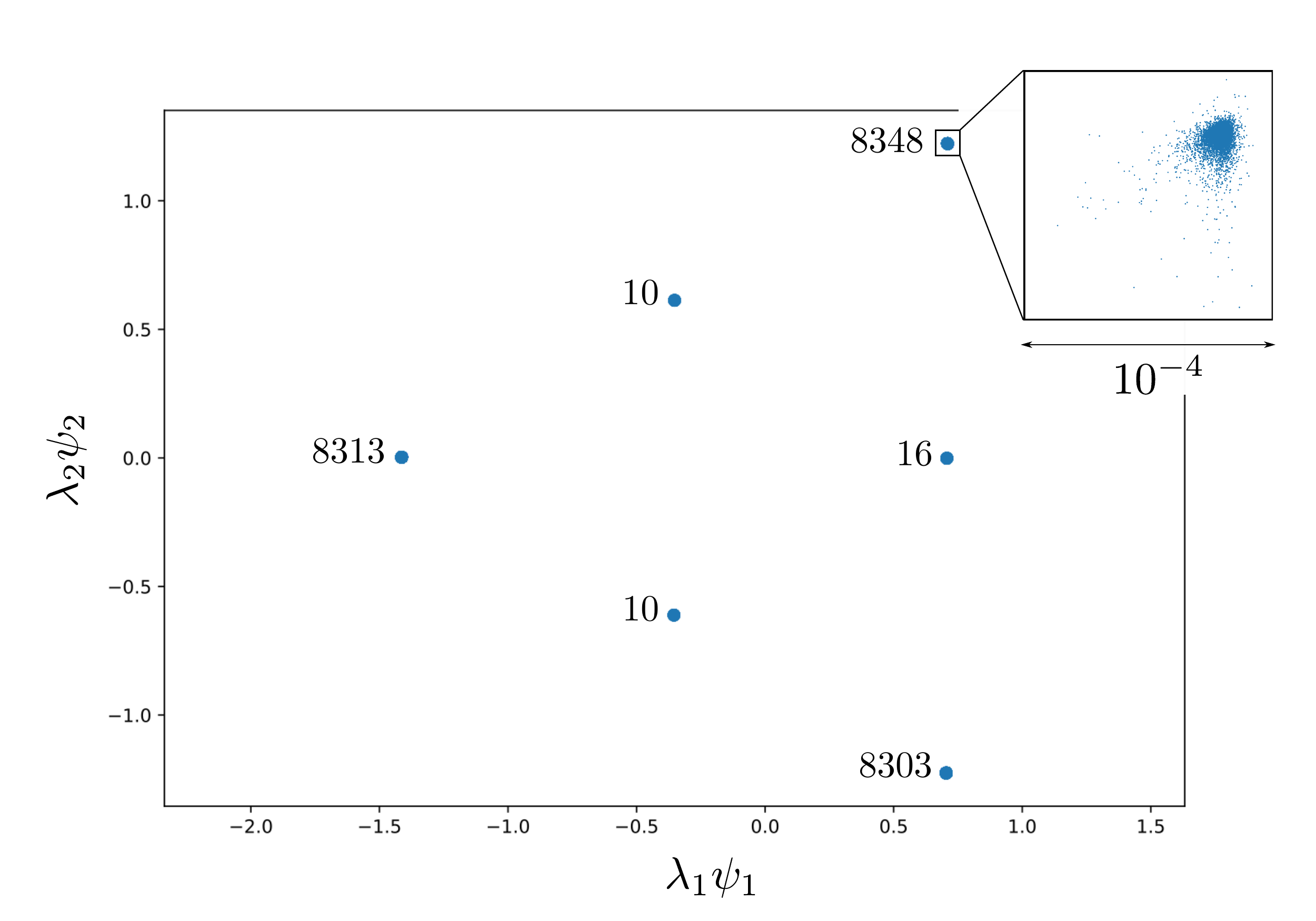}
\caption{Projection of the mess3's causal states on the reduced basis
	$\left\{ \Phi_{1},\Phi_{2}\right\}$. The number of samples stacked on
	each point is indicated.
	}
\label{fig:mess3proj}
\end{figure}

The eigenvalue spectrum is $\lambda_{0}=1$, $\lambda_{1}=0.999$,
$\lambda_{2}=0.999$ and $\lambda_{i\ge3}<1.2\times10^{-15}$. The algorithm thus
finds only two components, of equal significance. Figure~\ref{fig:mess3proj}
shows a scatter plot of the inferred causal states using coordinates in the
reduced basis $\left\{ \Phi_{1},\Phi_{2}\right\}$. The states are very well
clustered on the main vertices of the Sierpinski Gasket. (The inset shows that
$4$ orders of magnitude are needed to see a spread within each cluster,
compared to the main triangle size.) The triangle is remarkably equilateral and
its center even lies on $\approx(0,0)$, reflecting the symmetry of the original
HMM. To our knowledge, no other algorithm is currently able to correctly
recover the causal states purely from data from such a challenging, complex
process.

The number of samples at each triangle node is indicated. The nodes at the
first subdivision are about $800$ times less populated than the main nodes.
While theory predicts that states appear on all subdivisions of the Sierpinski
Gasket, the sample size needed to observe enough such states is out of reach
computationally.

\begin{figure*}
\includegraphics[width=.9\textwidth]{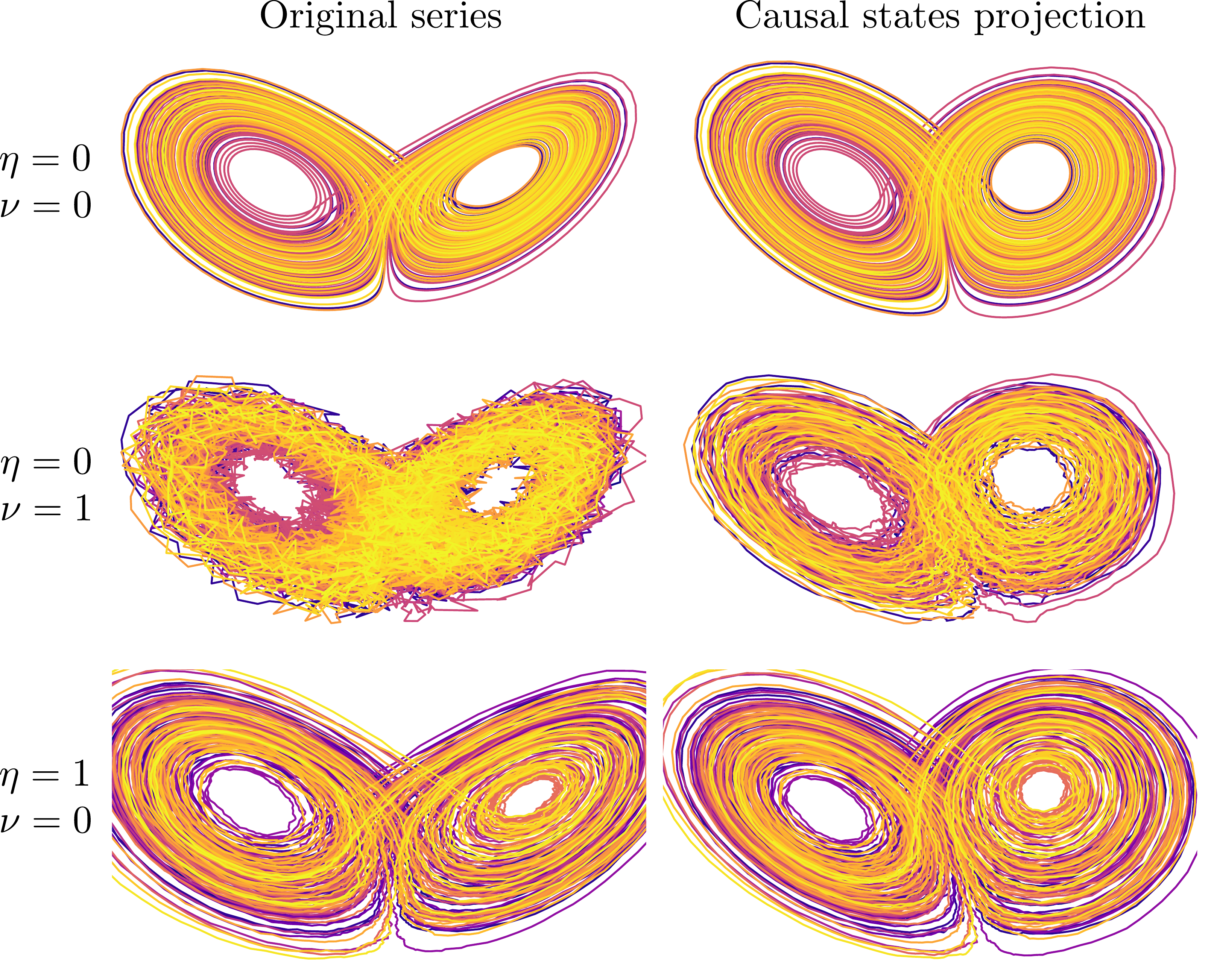}
\caption{Lorenz attractor (left) and its estimated \keM (right) at various
	thermal $\eta$ and measurement $\nu$ noise levels.
	}
\label{fig:Lorenz-attractor}
\end{figure*}

\subsection{Thermally-driven Continuous Processes: Chaotic Lorenz attractors}
\label{subsec:Lorenz-Attractor}

The first two examples demonstrated that \keM reconstruction works well on
discrete-valued data, even when states are expected to appear on a complicated
fractal structure or when the time series has infinite-range correlations. The
next step, and a central point of the development, is to reconstruct a
continuous infinity of causal states from sampled data. The following example
processes also serve to demonstrate time-series prediction using the estimated
\keM, recalling the method introduced in Sec.
\ref{subsec:Continuous_Trajectories}.

We first use the chaotic Lorenz ordinary differential equations from 1963 with
the usual parameters $\left(\sigma,\rho,\beta\right) = \left(10,28,8/3\right)$.
\cite{Lore63a} We add isotropic stochastic noise components $dW$ at amplitude
$\eta$ to model thermal fluctuations driving the three main macroscopic fluid
modes the ODEs were crafted to capture:
\begin{align*}
du & = -\sigma\left(u-v\right)dt+\eta dW\\
dv & = \left(\rho u-v-uw\right)dt+\eta dW\\
dw & = \left(-\beta w+uv\right)dt+\eta dW
  ~.
\end{align*}

A random initial condition $\left(u_{0},v_{0},w_{0}\right)$ is drawn in the
region near the attractor and a sufficiently long transient is discarded before
collecting data $\left(u_{t}, v_{t}, w_{t}\right)$. SDE integration is
performed using $dt=0.01$, yielding samples $\left(u_{t}, v_{t}, w_{t}
\right)_{0\leq t<T}$ up to a maximal time $T$ corresponding to $N=20,000$
(past, future) pairs.

In addition to the thermal fluctuations, we also model systematic measurement
error by adding a centered Gaussian process $\Gamma = \left(\gamma^{0},
\gamma^{1}, \gamma^{2}\right)$ with isotropic variance $\nu^{2}$. The result is
the observed series $\left(u'_{t}, v'_{t}, w'_{t}\right) =
\left(u_{t}+\gamma_{t}^{0}, v_{t}+\gamma_{t}^{1}, w_{t}+\gamma_{t}^{2}\right)$
used to estimate a \keM and perform a sensitivity analysis.

\subsubsection{\KeM reconstruction in the presence of noise and error}
\label{sec:keMReconstruction}

When $\eta=0$ and $\nu=0$ the deterministic ODEs' trajectories do not cross and
are uniquely determined by the initial condition $\left(u_t,v_t,w_t\right)$ at
$t=0$. Hence, each state on the attractor is its own causal state. Retaining
information from the past is moot, but only if $(u,v,w)$ is known with infinite
precision, due to the ODEs' chaotic solutions, that amplify fluctuations
exponentially fast. This is never the case in practice. So, considering small 
values of $\SeqLenX$ and $\SeqLenY$ may still be useful to better determine 
the causal states. We use $\SeqLenX=\SeqLenY=5$ in the reconstructions.

Figure~\ref{fig:Lorenz-attractor} shows the projections (right) with
coordinates $\left(\psi_{1},\psi_{2},\psi_{3}\right)$, together with the
original (pre-reconstruction) attractor data (left) for different noise
combinations.  Figure \ref{fig:Lorenz-attractor}(top row) displays the results
of \keM estimation from the noiseless data ($\nu = 0$ and $\eta = 0$).  The
second row shows the effects of pure measurement noise ($\nu = 1$ and  $\eta =
0$) on the raw data (left) and on the estimated \keM (right). Similarly, the
last row shows the effect of pure thermal noise ($\nu = 0$ and $\eta = 1$ .)

As expected, the structure is well recovered when no noise is added. A slight
distortion is observed. In practice, the causal states are unaffected by coordinate transforms and reparametrizations of $X$ and $Y$ which do not change equivalence in conditional distributions $P\left(Y|X=x_1\right)\equiv P\left(Y|X=x_2\right)$, so the algorithm could very well have found another parametrization of the Lorenz-63 attractor. See also Section \ref{sec:high-dim-lorenz96}. The causal states of the original data series are also well recovered even when that series is severely corrupted by strong measurement noise at $\nu=1$
in Fig.~\ref{fig:Lorenz-attractor}(middle row).

To appreciate these results recall that, in the noiseless case, if $x_{1}$ and
$x_{2}$ are in the same causal state $x_{2} \in \Eps\left(x_{1}\right)$ of the
original series, then by definition $\Pr\left(Y|X=x_{1}\right) =
\Pr\left(Y|X=x_{2}\right)$.
Measurement noise ($\nu>0$), independent at each time step, does not change
this. Since measurement noise is added to each and every time step
independently, noisy series $x'$ ending with the same noisy triplet
$\left(u',v',w'\right)$ at the current time $t=t_{0}$ end up in the same causal
state of the noisy system.  This is reduced to the current triplet
$\left(u',v',w'\right)$, itself a specific causal state of the deterministic
system. Hence, the causal states of the deterministic ODEs are subsets of
those of \keM estimated from the noisy-measured series. We arrive at the
important conclusion that, at least for deterministic chaotic systems, the
uniqueness and continuity of ODE solutions guarantee that \emph{causal states
are unaffected by measurement noise}. This is generally not true for
many-to-one maps and other functional state-space transforms that merge
states. 

In contrast to measurement noise, thermal noise ($\eta>0$) modifies the
equations of motion and the resulting trajectories reflect the accumulated
perturbations. Since each state on the (deterministic) attractor is
its own state, the estimated causal states are modified.

\begin{figure}
\includegraphics[width=1\columnwidth]{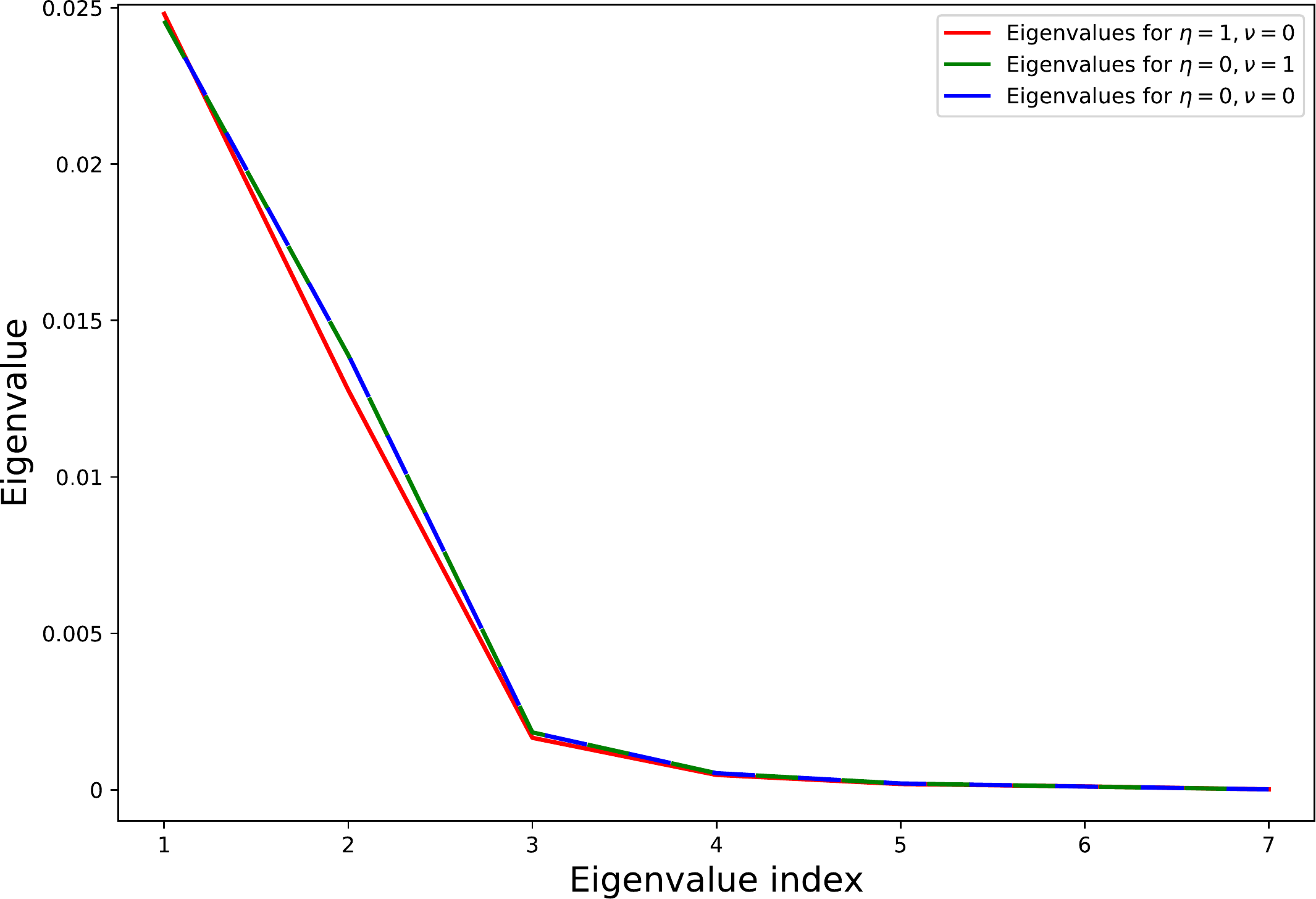}
\caption{Eigenvalues for the Lorenz-63 attractor estimated \keMs at various
	thermal and measurement noise levels.
	}
\label{fig:Lorenz-eigenvalues}
\end{figure}

Let's probe the robustness of \keM estimation.  With the parameters detailed
below, we obtain an eigenvalue spectrum shown in
Fig.~\ref{fig:Lorenz-eigenvalues}. There is an
inflection point after the first three components. The eigenvalues are
remarkably insensitive to a strong measurement noise level $\nu=1$. They are
also very robust to the thermal noise $\eta=1$, which induces only some
minor eigenvalue changes.

Thus, \keM estimation achieves a form of denoising beneficial for random
dynamical systems. However, the reconstructed causal states reflect the thermal
noise induced by $\eta=1$, as can be seen in the fine details of the bottom row
of Fig.~\ref{fig:Lorenz-attractor}. Note that the algorithm is able to strongly
reduce the measurement noise, and do so even while the attractor is very
corrupted, while the apparently minor thermal noise is preserved.

\subsubsection{\KeM prediction and sensitivity}
\label{sec:keMPrediction}

In Ref. \onlinecite{berry2015nonparametric}, a prediction experiment is performed using
an evolution operator computed directly in $\left(u,v,w\right)$ space instead
of in $\CSSet$, as here; recall Sec. \ref{subsec:Making-predictions}.  That
study focuses on how the error due to a small perturbation propagates with the
number of prediction steps.

Here, we use a more typical approach to first learn the model---i.e., compute
the states, the basis $\Phi$, the evolution operator, ...---on a data set of $N
= 20,000$ samples. Then, however, we estimate the prediction error on a
completely separate ``test'' set. This data set is generated with the same
parameters as for the training set, but starting from another randomly drawn
initial condition. $P=100$ test samples are selected after the transient, each
separated by a large subsampling interval so as to reduce their correlation.
Unlike the examples in the previous sections, this produces an objective error
function---the \emph{prediction accuracy}---useful for cross-validating the
relevant meta-parameters.

Due to the computational cost involved with grid searches, we only
cross-validate the data kernel $k^{V}$ bandwidth on a reduced data set in
preliminary experiments. In Ref. \onlinecite{berry2015nonparametric}, an arbitrary
variance was chosen for the distribution used to project $\left(u,v,w\right)$
triplets onto the operator eigenbasis. We also cross-validate it to improve the
results, for a fair comparison with \keM reconstruction.

\begin{figure}
\includegraphics[width=1\columnwidth]{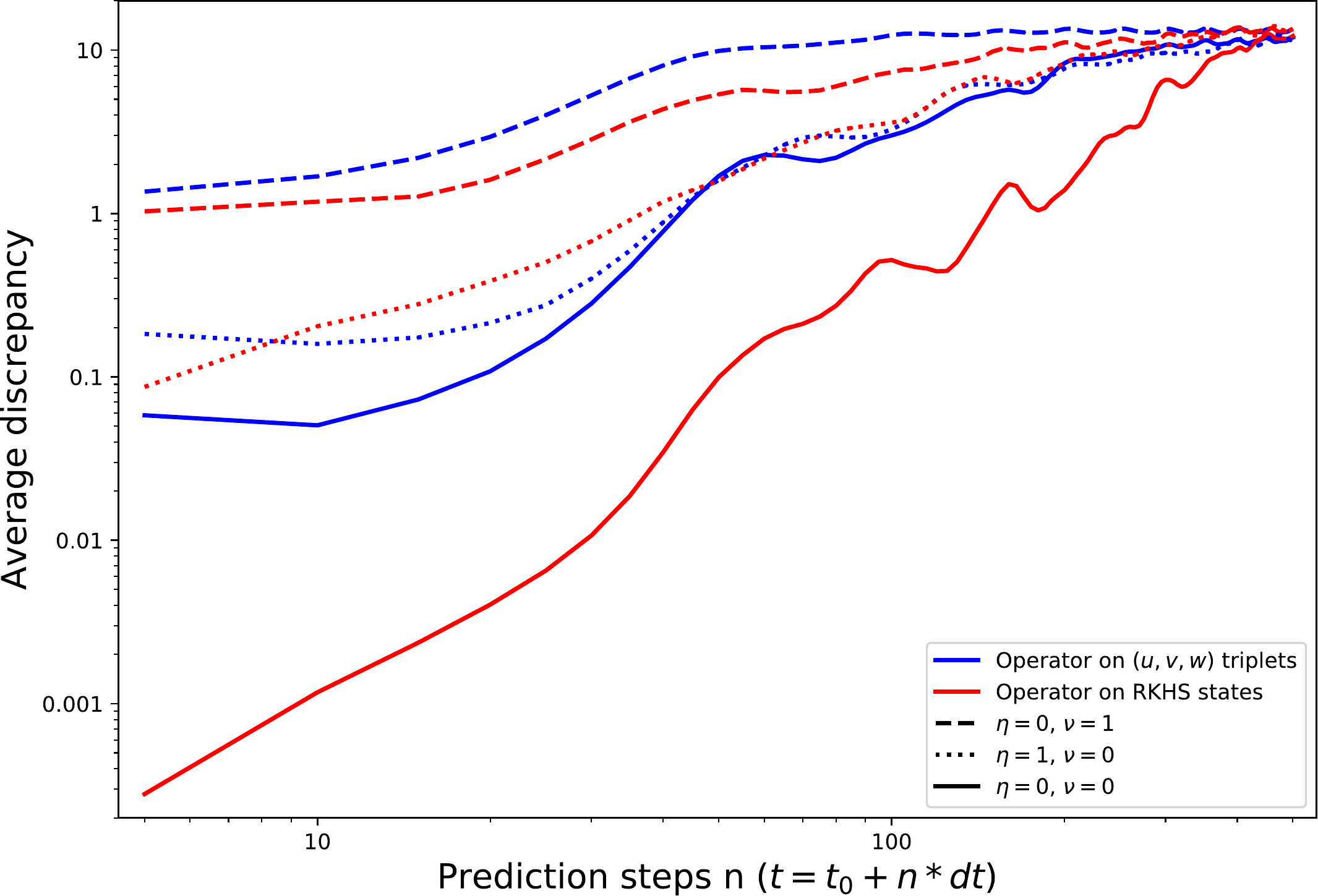}
\caption{Predictions on the Lorenz-63 attractor.
	}
\label{fig:Lorenz-predictions}
\end{figure}

Figure \ref{fig:Lorenz-predictions} presents the results. Forecasts are
produced at every $\Delta t=0.05$ interval and operator exponentiation is used
in between, as detailed in Sec.~\ref{subsec:Making-predictions}. This allows a
comparison of trajectories over $500$ elementary integration steps, which is
large enough for the trajectory to switch between the attractor main lobes
several times. Such trajectories are computed starting from each of the $100$
test samples.

In the noiseless case $\eta=\nu=0$, the average discrepancy $\left\langle
\left\Vert \left(u, v, w\right)-\left(u_{p}, v_{p}, w_{p}\right)\right\Vert
\right\rangle$ measures how predicted triplets $\left(u_{p}, v_{p},
w_{p}\right)$ differ from data triplets $\left(u, v, w\right)$, averaged at
each prediction step over the $100$ trajectories.

In the noisy cases, it make little sense to test the algorithm's ability to
reproduce noisy data. We instead test its ability to predict accurately the
noiseless series $\left(u, v, w\right)_{t>t_0}$ based on noisy past
observations $\left(u', v', w'\right)_{t\leq t_0}$. This is easily done for
measurement noise $\nu>0$, for which the original noiseless series is available
by construction.

For simulated thermal noise $\eta>0$, all we have is a particular realization
of an SDE, but no clean reference. Starting from the current noisy values at
each prediction point $\left(u', v', w'\right)$, we evolve a noiseless
trajectory with the basic Lorenz ODE equations. Since we use an isotropic
centered $\eta dW$ Wiener process, that trajectory is also the ensemble average
over many realizations of the SDE, starting from that same $\left(u', v',
w'\right)$ point. It makes more sense to predict that ensemble average, from
the current noisy causal-state estimate, rather than a particular SDE
trajectory realization.

The results in Fig.~\ref{fig:Lorenz-predictions} show a clear gain in precision
when using the RKHS method, both in the unperturbed data case and when data is
perturbed by measurement noise $\nu=1$. This gain persists until the trajectory
becomes completely uncorrelated with the original prediction point. The
situation is less favorable for thermal noise $\eta=1$. 

\begin{figure}
\includegraphics[width=1\columnwidth]{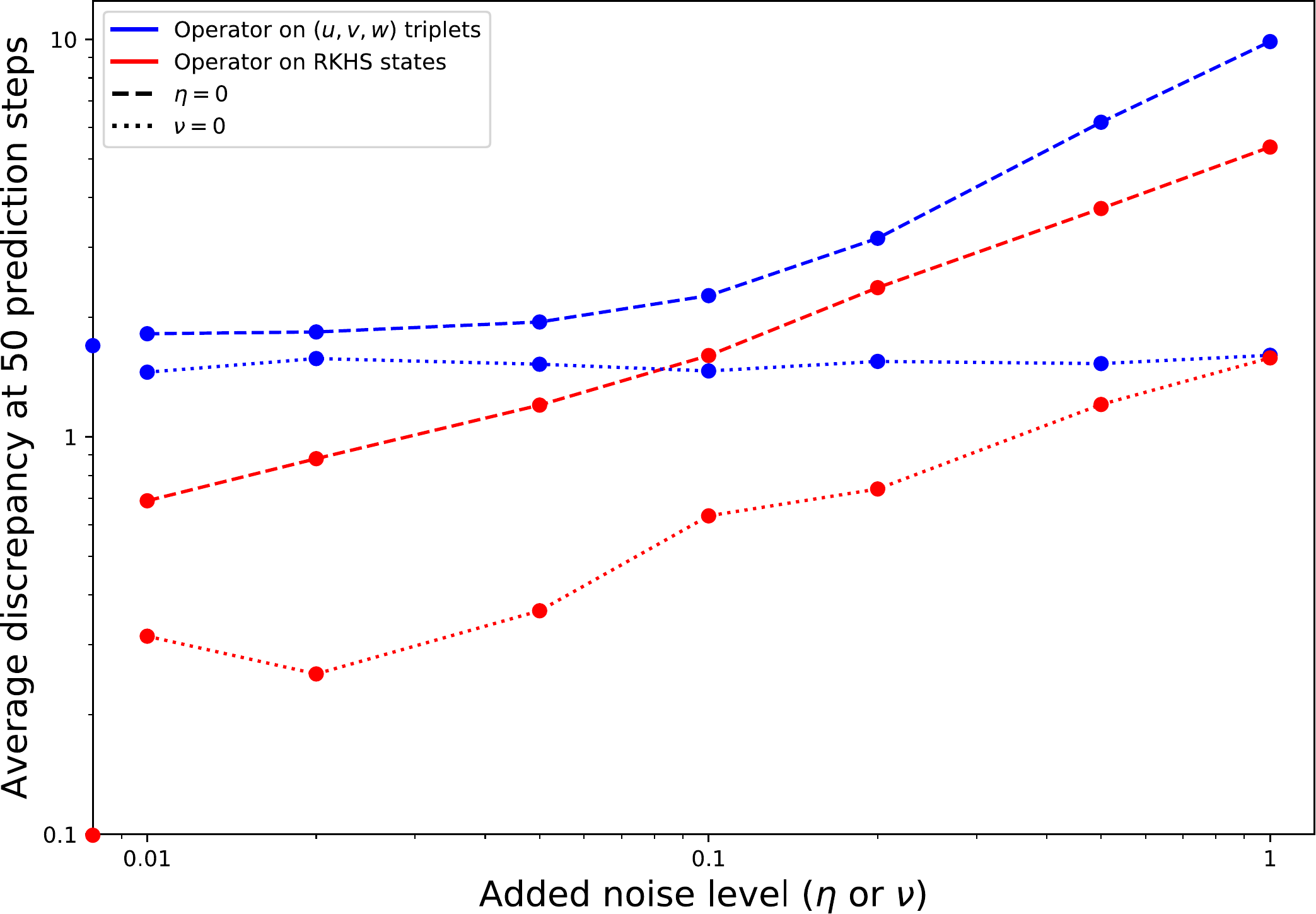}
\caption{Sensitivity when predicting trajectories along the Lorenz-63 attractor: Reconstruction error dependency on noise levels. Data for the $\nu=\eta=0$ noiseless case is shown as markers on the vertical axis.
}
\label{fig:Lorenz-noise-sensitivity}
\end{figure}

Figure~\ref{fig:Lorenz-noise-sensitivity} presents a sensitivity analysis that
focuses on predictions after $50$ time steps. For lower noise levels $\eta<1$,
the RKHS method still improves the prediction error, while the operator in data
space does not seem to be sensitive to $\eta$. We note (not shown here) that,
for longer time scales, the RKHS method may produce worse results on average.
The handling of measurement noise $\nu<1$ is also in favor of the RKHS method,
consistent with above results.

\begin{figure}
\includegraphics[width=1\columnwidth]{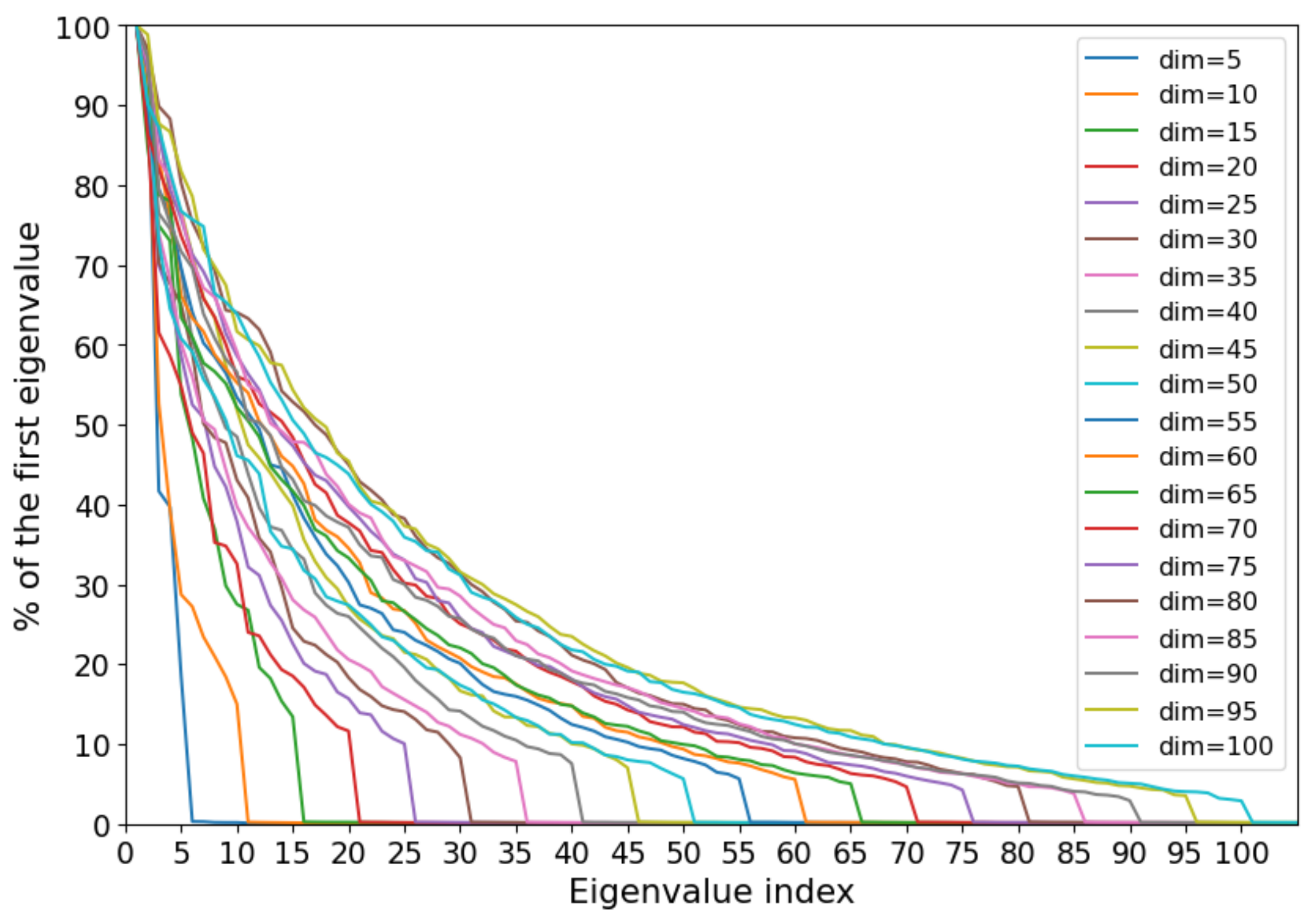}
\caption{Eigenvalues for the reconstruction of Lorenz-96 attractor, randomly projected in dimension 1000, with added high-dimensional noise, for $N=10000$ samples. The dimension in the legend refers to the original parametrization of Lorenz-96 attractor, which is recovered by the algorithm in the form of a spectral gap when reconstructed from the 1000-dimensional noisy time series. Values after the gap are very low and given in Table \ref{table:spectral_gaps}, together with the gap itself.}
\label{fig:Lorenz96-spectral-gaps}
\end{figure}

\begin{figure*}
\includegraphics[width=.99\textwidth]{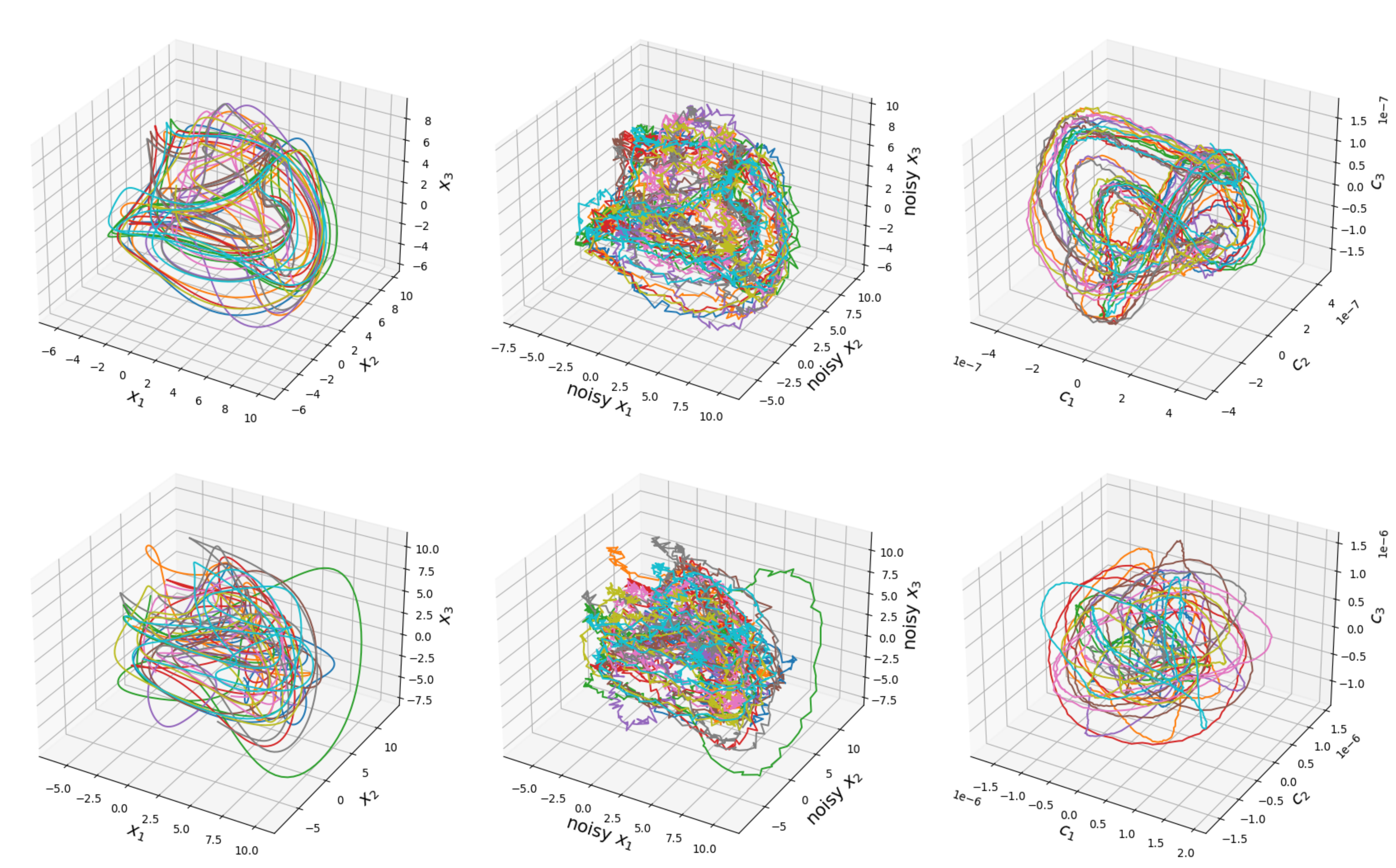}
\caption{(Top left) First 3 components of the Lorenz-96 $D=5$-dimensional
	attractor, with $N=5000$ samples. (Top middle) Same, back projected into
	the original space from the noisy 1000-dimensional random embedding. (Top
	right) First 3 coordinates of the causal states set $\CSSet$
	reconstruction. (Bottom row) same plots for the $D=100$-dimensional
	Lorenz-96 formulation. The reconstructed coordinates can be equivalent
	reparametrizations of the original variables and need not match those $1$
	to $1$. In each case the noise has been greatly reduced, as in Fig.
	\ref{fig:Lorenz-attractor}. Colors correspond to $10$ time series, each
	starting from a distinct random location on the basin of attraction, taken after a sufficiently long transient.
	}
\label{fig:Lorenz96-attractor}
\end{figure*}

\begin{table*}
\begin{tabular}{|c|c|c|c|c|c|c|c|c|c|c|c|c|c|c|c|c|c|c|c|c|c|c|c|c|c|}
\hline
Dim. $D$ & $5$ & $10$ & $15$ & $20$ & $25$ & $30$ & $35$ & $40$ & $45$ & $50$ & $55$ & $60$ & $65$ & $70$ & $75$ & $80$ & $85$ & $90$ & $95$ & $100$ \\
\hline
$100\frac{\lambda_{D+1}}{\lambda_1}$ & 0.31 & 0.23 & 0.26 & 0.24 & 0.24 & 0.24 & 0.22 & 0.25 & 0.26 & 0.21 & 0.23 & 0.23 & 0.23 & 0.22 & 0.23 & 0.26 & 0.24 & 0.20 & 0.25 & 0.21 \\
\hline
$\frac{\lambda_{D}}{\lambda_{D+1}}$ & 60.3 & 64.1 & 51.2 & 49.3 & 42.2 & 35.2 & 34.9 & 30.8 & 26.8 & 26.4 & 25.1 & 24.1 & 22.0 & 20.7 & 18.5 & 18.1 & 15.8 & 14.3 & 14.5 & 13.9 \\
\hline
\end{tabular}
\caption{Top row: Dimension of the Lorenz-96 ODE system; that is, before
	trajectories on the attractor undergo random projection and adding noise.
	Middle row: Eigenvalue after the spectral gap, expressed in \% of the first
	eigenvalue. Bottom row: Spectral gaps, relative to the value after the gap.
	Results for $N=10000$ samples are graphically presented in Fig.
	\ref{fig:Lorenz96-spectral-gaps}.
	}
\label{table:spectral_gaps}
\end{table*}

\begin{table}
\begin{tabular}{|c|c|c|c|c|c|c|c|c|c|}
\cline{2-10} \cline{3-10} \cline{4-10} \cline{5-10} \cline{6-10} \cline{7-10} \cline{8-10} \cline{9-10} \cline{10-10} 
\multicolumn{1}{c|}{} & \multicolumn{3}{c|}{$N=5000$} & \multicolumn{3}{c|}{$N=10000$} & \multicolumn{3}{c|}{$N=20000$}\tabularnewline
\hline 
Dim. $D$ & $90$ & $95$ & $100$ & $90$ & $95$ & $100$ & $90$ & $95$ & $100$\tabularnewline
\hline 
$100\frac{\lambda_{D+1}}{\lambda_{1}}$ & $0.22$ & $0.18$ & $0.21$ & $0.20$ & $0.25$ & $0.21$ & $0.21$ & $0.20$ & $0.21$\tabularnewline
\hline 
$\frac{\lambda_{D}}{\lambda_{D+1}}$ & $6.3$ & $5.8$ & $5.0$ & $14.3$ & $14.5$ & $13.9$ & $26.1$ & $25.6$ & $26.5$\tabularnewline
\hline 
\end{tabular}
\caption{Spectral gap dependency on the number $N$ of samples: Eigenvalues
	after the gap do not change in magnitude. Spectral gaps themselves, though,
	are much better resolved for larger $N$, especially in higher dimensions
	$D$.
	}
\label{table:gaps_dependency_on_N}
\end{table}

\subsection{Behavior with high-dimensional data and attractors}
\label{sec:high-dim-lorenz96}

This is all well and good, but do attractor reconstruction and denoising
capabilities hold in higher dimensions? In fact, it has been known for decades
\cite{Crut87a} that the Lorenz-63 system is special and very easily
reconstructed. This is due to its high state-space volume contraction rate and
its simple and smooth vectorfield that sports only two nonlinear terms.

To address these issues, we employ the Lorenz 1996 model,
\cite{Lorenz96} with tuneable dimension parameter $D$, defined for $i=1\ldots
D$ by $du_{i} / dt = -u_{i-2}u_{i-1}+u_{i-1}u_{i+1}-u_{i}+F$, with modulo
arithmetic on the indices. We use $F=8$, which yields chaotic dynamics. In
order to better cover all of the attractor, series are generated from $10$
different starting points taken at random on the bassin of attraction, after
discarding a sufficiently long transient. $N$ samples of $\left(x,y\right)$ 
pairs are collected from these series, using history lengths of 
$\SeqLenX=\SeqLenY=5$ values of vectors $u$. The projection of these series 
along the first 3 dimensions of the attractors for $D=5$ and $D=100$ are 
shown on the left part of Fig.~\ref{fig:Lorenz96-attractor}.

To test the algorithm's reconstruction performance, we embed the time series in
a $1000$-dimensional space using random projections: $V=UR$ with $R$ a matrix
of random components of size $D \times 1000$, taken from a normal distribution
of standard deviation $1/D$. $U$ holds the collected samples, organized in a
matrix of size $N\times D$. In practice, in such high dimension, the projection
directions in $R$ are nearly orthogonal. In addition, Gaussian noise with
variance $\nu^2=1/D$ is added to each component of $V$, similar to the setup in
Section \ref{sec:keMReconstruction}. We use the resulting data
$W=V+\textrm{noise}$ as input to the algorithm: Truly 1000-dimensional data,
now leaking into all dimensions thanks to the added noise, while retaining the
overall lower $D$-dimensional structure of the Lorenz-96 attractor. This
experiment seeks to reconstruct this hidden $D$-dimensional structure, solely
from noisy $1000$-dimensional time series.

The middle panes of Fig.~\ref{fig:Lorenz96-attractor} show the corrupted $W$
data, projected back into the original space using the pseudo-inverse
$pinv\left(R\right)$, for $D=5$ and $D=100$. The scaling of the noise variance
makes it so the strength of the noise is similar irrespective of $D$. 

The right panel of Fig.~\ref{fig:Lorenz96-attractor} shows the
well-reconstructed attractor. Moreover, the denoising for chaotic attractors
observed in Section \ref{sec:keMReconstruction} is also well reproduced in this
markedly more complicated setting. Interestingly, the first 3 coordinates of
the reconstructed attractor do not, and need not, match that of the original
$U$ series. Indeed, the causal states are equivalence classes of conditional
distributions and, as such, are invariant by reparametrizations of the original
data that preserve these equivalence classes; such as, coordinate
transformations.

We see that the algorithm finds a parametrization where each added coordinate
best encodes the conditional distributions in the low-dimensional coordinate
representation, as explained in Section \ref{subsec:Diffusion-maps}. Yet, that
parametrization encodes each and every initial $D$ component, even though it is
presented with $1000$-dimensional noisy time series of lengths $\SeqLenX=\SeqLenY=5$. This is shown in
Fig.~\ref{fig:Lorenz96-spectral-gaps} and Table \ref{table:spectral_gaps},
which clearly demonstrate spectral gaps at exactly $D$ reconstructed
components. These spectral gaps are more pronounced as the number of samples
$N$ increases, as shown in Table \ref{table:gaps_dependency_on_N}. As expected,
a larger number of samples $N$ is required for properly capturing the spectral
gaps as the dimension $D$ increases.

\section{Conclusion}

We introduced \keM reconstruction---a first-principles approach to empirically
discovering causal structure. The main step was to represent computational
mechanics' causal states in reproducing kernel Hilbert spaces. This gave a
mathematically-principled method for estimating optimal predictors of minimal
size and so for discovering causal structure in data series of wide-ranging
types.

Practically, it extends computational mechanics to nearly arbitrary data
types. (At least, those for which a characteristic reproducing kernel exists.)
Section \ref{subsec:RKHS_intro} showed, though, that this includes heterogeneous data types via kernel compositions.

Based on this, we presented theoretical arguments and analyzed cases for which
causal-state trajectories are continuous. In this setting, the \keM is
equivalent to an It{\^o} diffusion acting on the structured set of causal states---a
time-homogeneous, state-heterogeneous diffusion. The generator of that
diffusion and its evolution operator can be estimated directly from data. This
allows efficiently evolving causal-state distributions in a way similar to a
Fokker-Plank equation. This, in turn, facilitates predicting a process in its
original data space in a new way; one particularly suited to time series
analysis.

Future efforts will address the introduction of discontinuities, which may
arise for reasons mentioned in Sec. \ref{subsec:Causality-and-continuity}.
This will be necessary to properly handle cases where data sampling has occurred
above the scale at which time and causal-state trajectories can be considered
continuous. Similarly, when the characteristic scale of the observed system
dynamics is much larger than the sampling scale, a model reproducing the
dynamics of the sampled data may simply not be relevant. Extensions of the
current approach are thus needed, possibly incorporating jump components to
properly account for a measured system's dynamics at different scales. This,
of course, will bring us back to the computational mechanics of renewal and
semi-Markov processes.\cite{Marz17b}

Another future challenge is to extend \keM reconstruction to spatiotemporal
systems, where temporal evolution depends not only on past times but also on
spatially-nearby state values.\cite{Rupe17b,rupe2019disco,Rupe19a,Rupe20a} An
archetypal example of these systems is found with cellular automata. In fact,
any numerical finite elements simulation performed on discrete grids also falls 
into this category, including reaction-diffusion chemical oscillations and
hydrodynamic flows. Spacetime complicates the definition of the evolution
operator, compared to that of time series. However, the applicability of \keM
reconstruction would be greatly expanded to a large category of empirical data
sets.

Last, but not least, are the arenas of information thermodynamics and
stochastic thermodynamics. In short, it is time to translate recent results
on information engines and their nonequilibrium thermodynamics
\citep{Crut08b,Crut08a,Riec18b,Riec19a} to this broadened use of computational
mechanics. This, in addition to stimulating theoretical advances, has great
potential for providing new and powerful tools for analyzing complex physical
and biological systems, not only their dynamics and statistics, but also their
energy use and dissipation.

\section*{Data and code availability}
\label{sec:data availability}

The source code for the method described in this document is provided as
free/libre software and is available from this page:
\url{https://team.inria.fr/comcausa/continuous-causal-states/}. Experiments in
sections \ref{subsec:The-Even-Process}, \ref{subsec:Mixed-states} and
\ref{subsec:Lorenz-Attractor} can be reproduced by retrieving the tagged
version ``first\_arxiv'' from the GIT archive, the experiment in Section
\ref{sec:high-dim-lorenz96} with the tagged version ``chaos\_submission''.

\section*{Acknowledgments}
\label{sec:acknowledgments}

The authors thank Alexandra Jurgens for providing the code used for generating
samples from the ``mess3'' machine used in \ref{subsec:Mixed-states}. We also
thank Tyrus Berry for useful exchanges over nonlinear dimension reduction
through diffusion map variants and operator estimation methods
\citep{berry2015nonparametric,berry16local_kernels,berry16vbdm}. The authors
acknowledge the kind hospitality of the Institute for Advanced Study at the
University of Amsterdam. This material is based upon work supported by, or in
part by, Inria's CONCAUST exploratory action, Foundational Questions Institute
grant number FQXi-RFP-IPW-1902, U.S. Army Research Laboratory and the U.S.
Army Research Office grant W911NF-18-1-0028, and the U.S. Department of Energy under grant DE-SC0017324.

\section*{Reference}

\bibliographystyle{unsrt}

\end{document}